%% file: main.tex
\documentclass[10pt,twocolumn,letterpaper]{article}

\usepackage[pagenumbers]{cvpr}

\input{preamble}
\definecolor{linkblue}{rgb}{0.21,0.49,0.74}
\usepackage[pagebackref,breaklinks,colorlinks,allcolors=linkblue]{hyperref}

\title{Positive, Negative, or Unmentioned: Constrained 3D CT Report Generation via Masked Diffusion}

\author{
Dongchen Li$^{1}$\quad
Jitao Liang$^{1}$\quad
Yucheng Chen$^{2}$\quad
Xu Wang$^{1}$\quad
Guangliang Ju$^{3}$\quad
Wei Li$^{4,5}$\thanks{Corresponding author.}\\[6pt]
\normalsize
\begin{tabular}{c}
$^{1}$College of Computer Science and Engineering, Northeastern University, Shenyang, China\\
$^{2}$MedVisAI Lab, Lee Kong Chian School of Medicine, Nanyang Technological University, Singapore\\
$^{3}$Neusoft Medical Systems Co., Ltd., Shenyang, China\\
$^{4}$National Frontiers Science Center for Industrial Intelligence and Systems Optimization, Shenyang, China\\
$^{5}$Key Laboratory of Intelligent Computing in Medical Image (MIIC), Northeastern University, Shenyang, China\\[4pt]
{\tt\small \{2490254, 2472127\}@stu.neu.edu.cn, wangxu12@mails.neu.edu.cn, liwei@cse.neu.edu.cn,}\\
{\tt\small yucheng005@e.ntu.edu.sg, jugl@neusoftmedical.com}
\end{tabular}
}

\hypersetup{pdftitle={Positive, Negative, or Unmentioned: Constrained 3D CT Report Generation via Masked Diffusion},
            pdfauthor={Dongchen Li, Jitao Liang, Yucheng Chen, Xu Wang, Guangliang Ju, Wei Li}}

\begin{document}
\maketitle
\input{sec/0_abstract}
\input{sec/1_intro}
\input{sec/2_related}

\input{sec/3_method}
\input{sec/4_experiments}
\input{sec/5_conclusion}
{
    \small
    \bibliographystyle{ieeenat_fullname}
    \bibliography{refs}
}

\input{sec/X_suppl}

\end{document}

%% file: preamble.tex
\usepackage{xspace}

\usepackage{amssymb}
\usepackage{dsfont}  %
\usepackage{multirow}
\usepackage{float}  %
\usepackage{pifont}
\newcommand{\xmark}{\ding{55}}
\newcommand{\method}{ConDiff\xspace}
\newcommand{\planname}{Three-State Constraint Planning\xspace}
\newcommand{\assignname}{Latent Sentence Assignment\xspace}
\newcommand{\diffname}{Position-Constrained Canvas Diffusion\xspace}
\newcommand{\Write}{\textsc{Write}\xspace}
\newcommand{\Cov}{\textsc{Covered}\xspace}
\newcommand{\Skip}{\textsc{Skip}\xspace}
\newcommand{\eos}{\texttt{[EOS]}\xspace}
\newcommand{\void}{\texttt{[VOID]}\xspace}
\newcommand{\mask}{\texttt{[MASK]}\xspace}
\newcommand{\covtok}{\texttt{[COVER]}\xspace}

%% file: sec/0_abstract.tex
\begin{abstract}
CT reports state each observation as present or absent, or leave it unmentioned.
Radiologists report present findings but state an absent observation only when its absence answers the clinical question; in automatic generation, this choice of what to state becomes a constraint on each observation.
However, existing constraints for CT report generation are at most binary, with no unmentioned state, which leads to false negations of present findings.
They are also coarse, acting on a whole report or section, which leads to missed positive findings.
Masked diffusion language models (MDLMs) fill a canvas of masked tokens in any order, so a constraint can be placed where its statement is written.
We present ConDiff, \textbf{Con}strained Masked \textbf{Diff}usion, to our knowledge the first MDLM for 3D CT report generation, with one canvas segment per constrained observation.
Three-State Constraint Planning adds the unmentioned state and plans a negative statement only when it is worth the risk of a false negation.
Since a reference sentence can state multiple observations, Latent Sentence Assignment learns which of its segments writes it.
Position-Constrained Canvas Diffusion places each constraint within its segment, so each statement is generated under its own constraint.
On RadGenome-ChestCT and CTRG-Chest, ConDiff is state-of-the-art on almost all language, clinical, and polarity metrics, with the fewest false negations and the highest positive-finding F1.
Given the same constraints as a text prompt, an autoregressive model scores significantly lower than ConDiff on GREEN and RaTEScore.
Code will be publicly released.
\end{abstract}

%% file: sec/1_intro.tex
\section{Introduction}
\label{sec:intro}

\begin{figure}[t]
  \centering
  \includegraphics[width=0.9\linewidth]{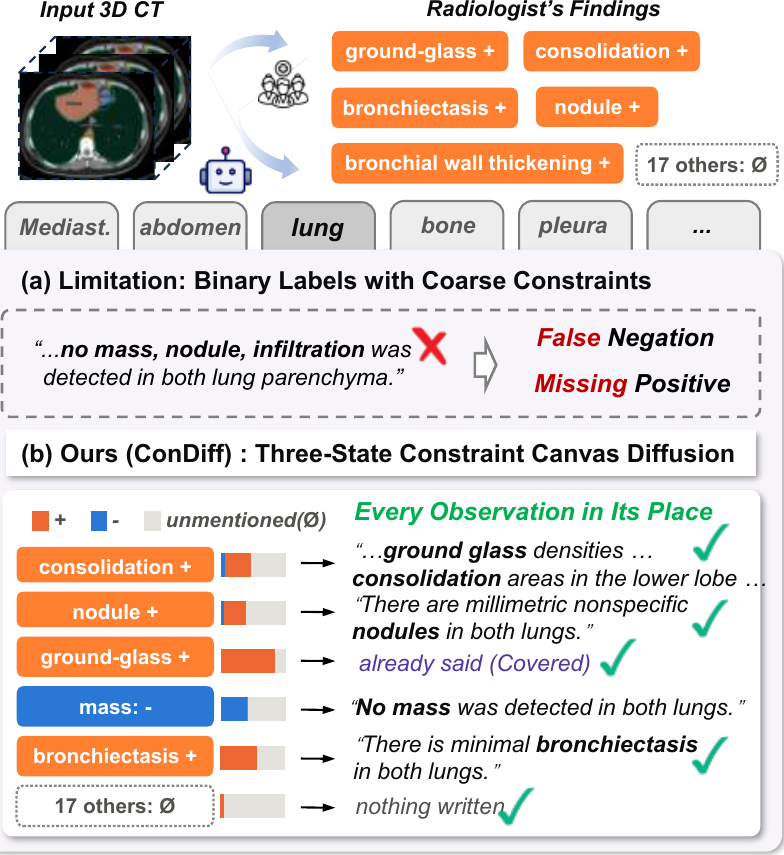}
  \caption{(a)~With binary labels and coarse constraints, a prior method's lung section states there is no nodule, a false negation, and misses the other positive findings. (b)~\method sets a three-state constraint per observation and gives each planned observation its own segment, stating four of five findings and one negative.}
  \label{fig:teaser}
\end{figure}

Automatic 3D chest CT report generation~\cite{ct2rep,ctrate,radgenome,reg2rg} drafts a radiology report from a chest CT scan.
A report states each observation as present or absent, or leaves it unmentioned~\cite{chexpert}, and 71\% of the statements in the reference reports of RadGenome-ChestCT~\cite{radgenome} are negatives such as ``no pleural effusion''.
Radiologists report present findings, but among absent observations they state only the pertinent negatives that answer the clinical question~\cite{esr2011,hartung2020} and leave the rest unmentioned.
A negative stated for a present finding is a \emph{false negation}~\cite{rexerr} that can delay diagnosis~\cite{bruno2015}; clinical error taxonomies rate it as a high-risk error, whereas leaving an absent observation unmentioned is a low-risk error~\cite{guan2025}.
Before writing, a radiologist therefore decides which observations are present and which absences are worth stating, and then describes them in precise radiologic language~\cite{hartung2020}.

When a report is generated automatically, this decision becomes a constraint on generation, with each observation reported in its decided state or left unmentioned.
Most methods encode the CT scan into visual features and generate the report with an autoregressive (AR) language model~\cite{ct2rep,ctrate,radfm,m3d,reg2rg}.
Recent methods further condition the language model on region features~\cite{reg2rg}, an organ graph~\cite{ctgraph}, differences from reference scans~\cite{diffvp}, or prompts.
Others first predict findings with a classifier~\cite{dialama,ctagrg,clargen}, since classifiers detect findings better than report generators do~\cite{promptmrg}.
Masked diffusion language models (MDLMs)~\cite{llada} have also been applied to report generation~\cite{anchordiff,echo,drrg}.
Despite this progress, existing methods do not constrain each observation as a radiologist does.

As \cref{fig:teaser}a shows, for a CT scan with consolidation and nodules, a prior method's lung section states that there is no nodule and misses the other positive findings.
This behavior stems from two limitations.
First, most methods set no explicit constraint, whereas classifier-guided CT report generators use binary labels that describe the image rather than the report, with no unmentioned state.
They thus do not decide which negatives to state: either the language model writes negatives on its own, or it follows the labels and states every undetected observation as absent.
As a result, prior methods with released predictions~\cite{reg2rg,diffvp,ctgraph} on the RadGenome-ChestCT validation set make 33 to 64 false negations per 100 reports, as \cref{tab:sota} shows.
Second, existing constraints act on a whole report or region section, without specifying which sentence reports each observation and its state.
Specifying this assignment is not straightforward, because real reports often state multiple observations in one sentence, such as ``no pericardial effusion or thickening''.
As a result, the model can miss positive findings that the constraints specify, and prior methods with released predictions recover only 11\% to 25\% of the positive findings in the reference reports.
Hence, existing constraints neither match what a report states nor act where each statement is written.

To address these problems, we build on MDLMs, which attend in both directions and fill a canvas of masked tokens in any order, so a constraint can be placed where its statement is written.
Unlike existing MDLM methods, which write a whole report or section on a single canvas, we set a constraint for each observation and place it where the observation is written.
First, each observation receives a three-state constraint, and an expected-cost rule decides which negatives are worth stating while leaving the remaining observations unmentioned.
Second, each observation to be stated has its own segment, whose first position decides whether the segment is written before any word is generated.
Every written segment is thus generated under the constraint of its own observation.
We thus present ConDiff, \textbf{Con}strained Masked \textbf{Diff}usion, for 3D CT report generation.
To our knowledge, it is the first MDLM for this task, and it uses one canvas per anatomical region and one segment per planned observation.

In \method, \planname sets each observation's constraint from three-state classifiers using the expected-cost rule.
When a reference sentence satisfies multiple constraints, \assignname treats the segment that writes it as a latent variable and learns this assignment via hard expectation maximization (EM), while the segments of its other observations are marked as \emph{covered}.
This assignment keeps generated sentences close to those in real reports.
\diffname places each constraint within its segment via a prototype of its observation and state, and the first position of a segment decides whether the segment is written or covered, without requiring a tunable threshold.
On generated reports, $N$-gram metrics, label-based clinical efficacy, GREEN~\cite{green}, and RaTEScore~\cite{ratescore} do not penalize a false negation more heavily than leaving the present finding unmentioned.
We therefore also report polarity metrics that score positives, negatives, and false negations separately.
We evaluate \method with both standard and polarity metrics on two 3D chest CT benchmarks.

Our contributions are:
\begin{itemize}
  \item We present \method, to our knowledge the first MDLM for 3D CT report generation, where \assignname learns by hard EM which segment writes each sentence and \diffname places each constraint within its segment.
  \item We propose \planname, which adds the unmentioned state and plans a negative statement only when it is worth the risk of a false negation.
  \item We report polarity metrics that score false negations separately, and on two benchmarks, \method is state of the art on almost all metrics and significantly outperforms an AR model on GREEN and RaTEScore under identical plans.
\end{itemize}

%% file: sec/2_related.tex
\section{Related Work}
\label{sec:related}

\paragraph{3D CT report generation.}
Most 3D CT report generators, such as CT2Rep~\cite{ct2rep}, CT-CHAT~\cite{ctrate}, RadFM~\cite{radfm}, M3D~\cite{m3d}, and 3D-CT-GPT~\cite{ctgpt3d}, pair a visual encoder with an autoregressive (AR) language model.
Later work improves the visual input with multi-view perception~\cite{mvketr}, hierarchical injection~\cite{uvlm}, contrastive alignment~\cite{liu2025structure,su2025contrastive,clearview}, stronger encoders~\cite{nvreasonct,btb3d}, or latent world-state tokens~\cite{sliceworld}.
Other methods condition the language model on region features~\cite{reg2rg}, an organ graph~\cite{ctgraph}, per-region agents~\cite{march}, entity prompts~\cite{dear}, differences from a normal reference scan~\cite{diffvp}, or retrieved reports and evidence~\cite{rafs,adaragct}.
Classifier-guided methods~\cite{dialama,ctagrg,clargen} and cue prompts~\cite{dcppd} supply binary present-or-absent labels, and Abn-BLIP~\cite{abnblip} uses abnormality queries for CT pulmonary angiography.
On chest X-rays, PromptMRG~\cite{promptmrg} adds a blank class to its prompts but takes the most likely class of each disease.
CLarGen~\cite{clargen} and AdaRAG-CT~\cite{adaragct} observe that generated reports drift toward generic normal template sentences.
Evaluation relies on $n$-gram metrics, label-based clinical efficacy~\cite{chexpert,ctrate}, RadGraph F1~\cite{radgraph,radgraphf1}, GREEN~\cite{green}, and RaTEScore~\cite{ratescore}.
CT-FineBench~\cite{ctfinebench} scores the attributes of present findings and ignores negation errors, and ReRef~\cite{reref} shows on chest X-rays that compressing the normal statements in reference reports changes model rankings.
However, existing constraints for CT report generation are at most binary, with no unmentioned state, and act on a whole report or region section.

\paragraph{Masked diffusion language models for radiology.}
Discrete diffusion models such as D3PM~\cite{d3pm}, SEDD~\cite{sedd}, and MDLM~\cite{mdlm} denoise a text canvas iteratively, and LLaDA~\cite{llada} and Dream~\cite{dream} scale MDLMs to the size of AR language models.
Instruction-tuned MDLMs pad a response to a fixed canvas length with end tokens, which triggers early termination~\cite{rainbow}, so Rainbow Padding~\cite{rainbow} uses a cycle of distinct padding tokens, VoidPadding~\cite{voidpadding} uses a dedicated \void token, and DAEDAL~\cite{daedal} adapts the canvas length.
In radiology, MDLMs have been used only for chest X-ray reports.
AnchorDiff~\cite{anchordiff} protects RadGraph entities during masking, DRRG~\cite{drrg} conditions on concepts with entity-aware masking, ECHO~\cite{echo} denoises a block in one step, and Van Puyvelde \etal~\cite{cxrdiffdraft} match an AR model of the same size under the same fine-tuning recipe on medical visual question answering and also draft reports by infilling.
These methods write a whole report or a single section on one canvas, without a constraint for each statement.

%% file: sec/3_method.tex
\section{Method}
\label{sec:method}

\begin{figure*}[t]
  \centering
  \includegraphics[width=\linewidth]{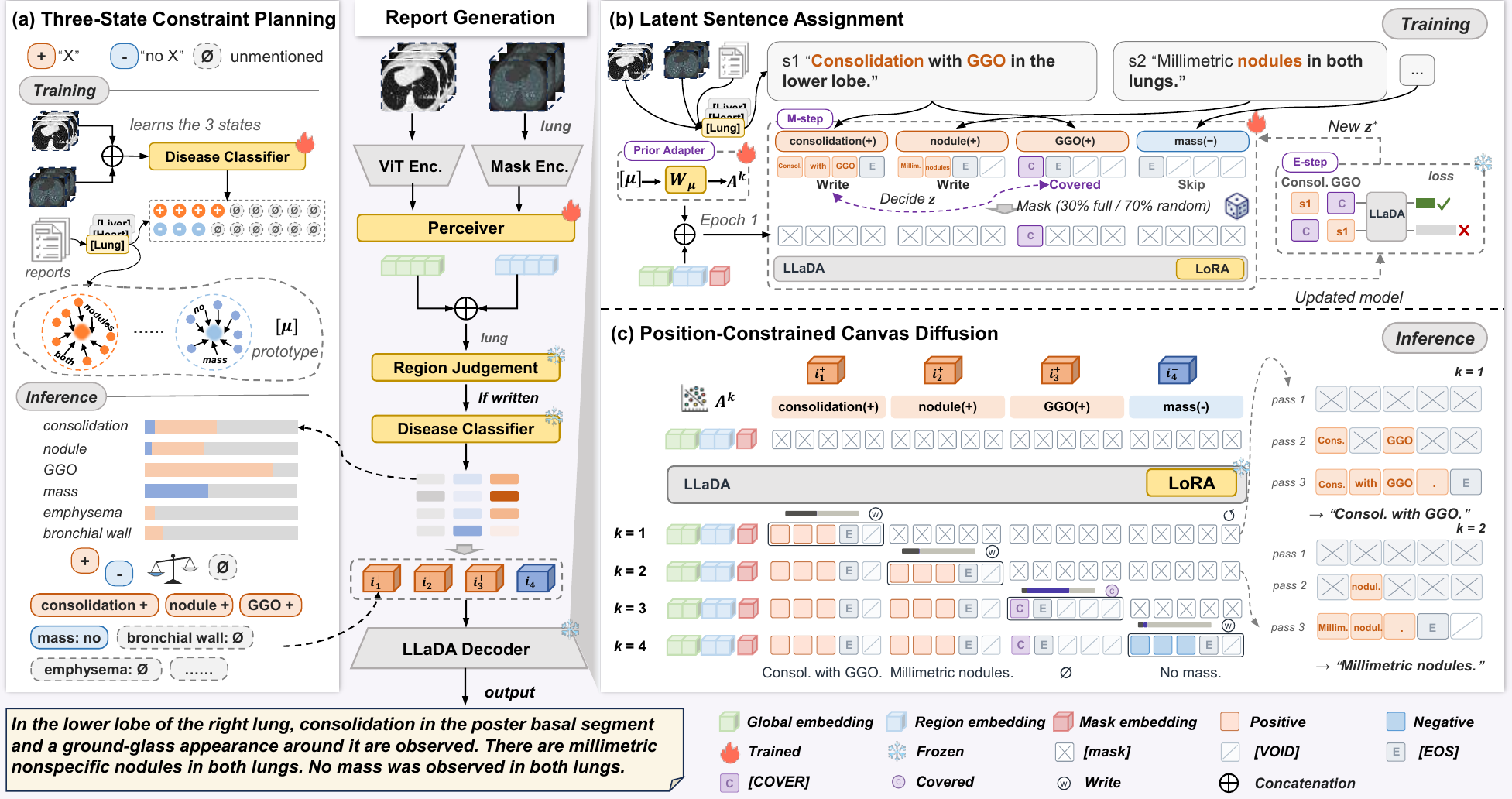}
  \caption{Overview of \method. (a)~\textbf{\planname} sets each observation's constraint by expected cost under its three-state classifier. (b)~\textbf{\assignname} learns by hard EM which segment writes a sentence that satisfies multiple constraints. (c)~\textbf{\diffname} places each constraint within its segment, whose first position chooses \Write or \Cov.}
  \label{fig:method}
\end{figure*}

\method writes a CT report one anatomical region at a time on an \emph{observation canvas} with one segment per planned observation (\cref{fig:method}).
After defining the observations and the canvas in \cref{sec:formulation}, we describe the three modules in \cref{sec:planning,sec:assignment,sec:diffusion}.

\subsection{Problem Formulation}
\label{sec:formulation}

\paragraph{Observation states.}
A CT volume $V$ comes with masks for a fixed set of regions $\mathcal{R}$, and its report has one section per region, each described by \emph{observations} in three states.
Following CheXpert~\cite{chexpert}, observations are (organ, finding category) pairs such as (pleura, effusion) or (lung, nodule) and normal statements such as ``the heart is of normal size''.
Large language models build the observation vocabulary $\mathcal{I}_r$ of region $r$ once from the training reports, as detailed in \cref{supp:method}.
A sentence $u$ states the observations $C(u) \subseteq \mathcal{I}_r$, where each observation $i$ has a negative or positive status $\sigma_u(i) \in \{1, 2\}$ and uncertain counts as positive.
The state of observation $i$ in a section $S$ is
\begin{equation}
  y_i(S) = \max\big(\{0\} \cup \{\sigma_u(i) : u \in S,\ i \in C(u)\}\big),
  \label{eq:state}
\end{equation}
where $0$ means unmentioned, $1$ stated negative, and $2$ stated positive, so ``no pleural effusion'' sets $y_{\text{pleura, effusion}} = 1$ and a section that never mentions effusion leaves it at $0$.
Observations for normal statements never take state $2$.
The observations of a region follow a canonical order given by the mean relative position of their sentences in the training sections.

\paragraph{Observation canvas.}
The observation canvas of a section has one segment per observation in its plan.
A plan $P_r = \big((i_1, s_1), \dots, (i_K, s_K)\big)$ lists the observations a section states in canonical order, each with state $s_k \in \{1, 2\}$.
A plan thus sets a three-state constraint for every observation, as each listed observation is stated in its state $s_k$ and every other observation is left unmentioned.
Segment $k$ has a fixed length $n_k$, set from the text length of observation $i_k$ predicted from the region feature and the planned state (\cref{sec:setup}).
It holds the text, an end-of-segment token \eos, and \void padding~\cite{voidpadding} to its end so every segment can terminate early, and the section joins segment texts in canvas order.

\paragraph{Prototypes.}
Each stated constraint $(i, s)$, an observation $i$ in state $s \in \{1, 2\}$, is represented by a \emph{prototype} $\boldsymbol{\mu}(i, s)$ computed once from the training reports as
\begin{equation}
  \boldsymbol{\mu}(i, s) = \frac{\sum_{u \in T_{i,s}} c_u\, \bar{\mathbf{E}}(u)}{\sum_{u \in T_{i,s}} c_u},
  \label{eq:proto}
\end{equation}
where $T_{i,s}$ holds the most frequent training sentences whose first observation in canonical order is $i$ in state $s$, $c_u$ counts sentence $u$, and $\bar{\mathbf{E}}(u)$ averages its frozen input token embeddings.
A prototype thus encodes an observation and its state in the fixed input space of the language model, so the observations of a new dataset need no learned embedding.

\paragraph{Masked diffusion.}
In masked diffusion, an MDLM~\cite{mdlm} such as LLaDA~\cite{llada} corrupts a sequence $x_0$ into $x_t$ by replacing each token with \mask independently with probability $t \sim \mathcal{U}(0, 1)$, and a bidirectional Transformer $p_\theta$ predicts every masked token of $x_t$ at once.
It is trained with
\begin{equation}
  \mathcal{L}_{\text{MDLM}}(\theta) = \mathbb{E}_{t, x_t}\Big[\frac{1}{t} \sum_{j:\, x_t^j = \text{\mask}} -\log p_\theta\big(x_0^j \mid x_t\big)\Big],
  \label{eq:mdlm}
\end{equation}
and generates by starting from a fully masked sequence and committing its most confident predictions over several forward passes.
On the observation canvas, every segment is visible to every other, and the first position of a segment holds a distribution over how the segment starts before any of its tokens are written.
\Cref{sec:assignment,sec:diffusion} use both properties.

\subsection{\planname}
\label{sec:planning}

\paragraph{Observation state probabilities.}
For each observation $i$, a multinomial logistic regression with an L2 penalty serves as a three-state classifier that maps a region feature $\mathbf{h}_r$, pooled from the frozen CT encoder tokens, to the observation state probabilities $\mathbf{p}_i = (p_{i0}, p_{i1}, p_{i2})$ using the states in \cref{eq:state} as labels.
On the training set, these probabilities are out-of-fold estimates, with folds split by patient so that no training CT is scored by a classifier fitted on its own report.
A logistic gate for each region, fitted the same way, decides whether to write the region section using the out-of-fold F1 optimum as its threshold.
We treat these probabilities as approximate posteriors below, and \cref{supp:method} shows that the positive share $p_{i2} / (p_{i1} + p_{i2})$ is well calibrated.

\paragraph{Choosing what to state.}
We choose what to state in two steps by assigning every observation $i$ an action $a_i$ of stating it as positive ($+$), stating it as negative ($-$), or leaving it unmentioned ($\varnothing$).
First, if stating ``$i$'' costs $\alpha_i$ relative to a missed finding when the reference does not state $i$ as present, stating it has lower expected cost when $p_{i2} > \eta_i = \alpha_i / (1 + \alpha_i)$, and we set $\eta_i$ directly to the out-of-fold F1 optimum among thresholds that meet a precision floor.
Second, leaving $i$ unmentioned costs $1$ when the reference states ``no $i$''.
Stating ``no $i$'' instead costs $\kappa$ when the reference leaves $i$ unmentioned, $0$ when it states the same negative, and $\lambda$ more than leaving $i$ unmentioned when it states $i$ as present, which makes the negative a false negation.
Comparing the expected costs gives the plan rule
\begin{equation}
  a_i = \begin{cases}
    + & \text{if } p_{i2} > \eta_i,\\
    - & \text{if } p_{i2} \le \eta_i \text{ and } p_{i1} > \lambda\, p_{i2} + \kappa\, p_{i0},\\
    \varnothing & \text{otherwise}.
  \end{cases}
  \label{eq:rule}
\end{equation}
A larger $\kappa$ leaves more observations unmentioned.
This cost ordering follows a clinical error taxonomy for report generation that rates a false negation as high risk and both an unstated negative and a negative lacking support in the reference as low risk~\cite{guan2025}.
At $\kappa = -1$, the second condition of the $-$ case becomes $p_{i2} < 1/(1 + \lambda)$, so every observation with $p_{i2}$ below both $\eta_i$ and $1/(1 + \lambda)$ is stated as negative, which approximates the standard two-state formulation.

\subsection{\assignname}
\label{sec:assignment}

\paragraph{Write, Covered, and Skip.}
Every canvas segment takes one of the three targets \Write, \Cov, or \Skip.
A sentence stating multiple planned observations must be written by exactly one of their segments without the others repeating it, even though the reference does not specify which segment writes it.
Let $C'(u) = C(u) \cap \{i_1, \dots, i_K\}$ be the canvas observations of sentence $u$, where we drop any sentence with none, let $z_u \in C'(u)$ be the observation whose segment writes $u$, and let $U_k(z) = \{u \mid z_u = i_k\}$ be the sentences assigned to segment $k$ in report order.
An assignment $z = (z_u)_u$ fixes the target $x_0^{(k)}(z)$ of every segment $k$ (\cref{fig:method}b).
If $U_k(z) \neq \emptyset$, the target is \Write, consisting of the text of $U_k(z)$ followed by \eos and \void.
If $U_k(z) = \emptyset$ but some sentence states $i_k$, the target is \Cov, consisting of \covtok followed by \eos and \void.
Otherwise the report leaves $i_k$ unmentioned, and the target is \Skip, consisting of only \eos and \void, which does not depend on $z$.
Because the three targets begin with distinct tokens, the first position of a segment selects among them before any text is written (\cref{sec:diffusion}).

\paragraph{Generator training.}
Generator training uses noisy reference plans as training canvases.
Each training canvas holds the observations of the reference section in their reference states, with each state flipped between negative and positive with a small probability so the generator does not follow the plan blindly.
For observations that the report does not state despite a high out-of-fold mention probability $1 - p_{i0}$, we add \Skip segments in their more likely stated state to teach the generator that a planned observation need not be written.
Given the targets $x_0(z)$ of an assignment $z$ on these canvases, we fine-tune the MDLM with LoRA~\cite{lora}, along with the resampler and $W_\mu$ of \cref{sec:diffusion} and a \void output adapter.
To match the start of decoding, where the whole canvas is masked, the corruption masks the whole canvas with probability $f$.
Otherwise, it masks text and \void positions with probability $t \sim \mathcal{U}(0, 1)$ and always masks \eos.
The loss is a cross-entropy weighted by position type,
\begin{equation}
  \mathcal{L}_{\text{gen}}(\theta) = \mathbb{E}\Big[\sum_{j \in \mathcal{M}} \bar{w}_j \big(-\log p_\theta(x_0^j \mid x_t, \mathbf{c}_r)\big)\Big],
  \label{eq:loss}
\end{equation}
where $\mathcal{M}$ is the set of masked positions in a batch, $\mathbf{c}_r$ is the visual prefix, and $\bar{w}_j$, normalized to sum to one over $\mathcal{M}$, is the weight of the type $\tau(j) \in \{\text{text}, \text{\eos}, \text{\void}\}$ of position $j$, where \covtok counts as text.
Unlike \cref{eq:mdlm}, this loss drops the $1/t$ weight and is therefore a weighted cross-entropy rather than a likelihood bound.
We write $\ell_t(z)$ for its unnormalized sum over one section with targets $x_0(z)$ at masking rate $t$.

\paragraph{Hard EM.}
We treat the assignment $z$ as a latent variable and learn it by hard EM~\cite{dempster1977},
\begin{equation}
  \min_\theta \sum_{(V, r)} \min_{z \in \mathcal{Z}_{V,r}} L(\theta; z), \qquad \mathcal{Z}_{V,r} = \prod_u C'(u),
  \label{eq:em}
\end{equation}
where $L(\theta; z)$ is the expected training loss of a section with targets $x_0(z)$ under the corruption above.
Unlike the alignments of CTC~\cite{ctc} or monotonic alignment search~\cite{glowtts}, the assignment need not follow sentence order, and \cref{supp:method} shows that many training sections admit no assignment that does.
The first epoch uses the first-observation assignment, which assigns each sentence to its first observation in canonical order.
Then the E-step fixes $\theta$ and selects the assignment with the lowest loss for every training section, and the M-step fixes $z$ and trains the generator on its targets for one epoch.
Because the corruption masks the whole canvas with probability $f$, the E-step approximates the expected loss as
\begin{equation}
  \begin{aligned}
    L(\theta; z) &= f\, \ell_1(z) + (1 - f)\, \mathbb{E}_t\, \ell_t(z) \\
         &\approx f\, \ell_1(z) + (1 - f)\, \ell_{1/2}(z) = \hat{L}(z).
  \end{aligned}
  \label{eq:estep}
\end{equation}
On the fully masked canvas, where decoding starts, a single forward pass computes $\ell_1$ exactly for every $z$ because the masked input holds only \mask tokens and plan embeddings and is independent of $z$.
Segment lengths do not depend on $z$ either, so positions correspond across candidates and $\ell_{1/2}$ shares a single mask draw across all candidates of a section.
Because a section has few assignments, the E-step enumerates them under $\ell_1$ and evaluates $\ell_{1/2}$ only on the top assignments and the first-observation assignment, so the selection is approximate.

\subsection{\diffname}
\label{sec:diffusion}

\paragraph{Visual prefix.}
The visual prefix conditions the generator on the volume and region.
It starts from a frozen NV-Reason-CT encoder~\cite{nvreasonct} that maps the volume to tokens on a 3D grid.
A shared Perceiver resampler~\cite{flamingo} turns the grid after average pooling into global tokens and the cells covered by the dilated region mask, each with an embedding of its coverage fraction, into region tokens.
Embeddings of the region mask and region name follow, and the canvas comes after this prefix.

\paragraph{Position constraints.}
Each constraint $(i_k, s_k)$ of the plan enters the generator only through the masked positions of its segment $k$, in training and at inference.
The input of canvas position $j$ in segment $k(j)$ is
\begin{equation}
  \mathbf{x}_j = \mathbf{E}(x_t^j) + \mathds{1}[x_t^j = \text{\mask}]\, W_\mu\, \boldsymbol{\mu}(i_{k(j)}, s_{k(j)}),
  \label{eq:position}
\end{equation}
where $\mathbf{E}$ is the input embedding of the language model and $W_\mu$ is learned and initialized to zero so that training begins from the language model without a plan.

\paragraph{Decoding.}
Segments are decoded sequentially in canvas order, and each forward pass sees the whole canvas.
Because the plan has already chosen to state the observation of segment $k$, its first position decides only between \Write and \Cov.
With repetitions and misses weighed equally, the segment is written when the total probability $q^{\text{write}}_k$ of text tokens exceeds the probability $q^{\text{cov}}_k$ of \covtok, with no tuned threshold.
The last segment of a section is always written if no earlier one was written.
\Cref{supp:method} details how a written segment is filled.

%% file: sec/4_experiments.tex
\begin{table*}[t]
  \caption{Results on RadGenome-ChestCT validation and CTRG-Chest test. Scores are $\times 100$ and FNeg is per 100 reports. MTR is METEOR and RaTE is RaTEScore. $^\dagger$From Reg2RG~\cite{reg2rg}. $^\ddagger$Rescored official predictions. $^\S$On its 1,505 released CTs.}
  \label{tab:sota}
  \centering
  \small
  \setlength{\tabcolsep}{4pt}
  \begin{tabular}{@{}l ccccc cccc ccc@{}}
    \toprule
    & \multicolumn{5}{c}{Language} & \multicolumn{4}{c}{Clinical} & \multicolumn{3}{c}{Polarity} \\
    \cmidrule(lr){2-6} \cmidrule(lr){7-10} \cmidrule(l){11-13}
    Method & B-1 & B-4 & MTR & R-L & CIDEr & CE-F1 & RadGraph & GREEN & RaTE & Pos-F1 & Neg-F1 & FNeg$\downarrow$ \\
    \midrule
    \multicolumn{13}{@{}l}{\emph{RadGenome-ChestCT}} \\
    R2GenGPT~\cite{r2gengpt}$^\dagger$ {\scriptsize(Meta-Radiol.'23)} & 43.3 & 24.2 & 39.9 & 32.3 & -- & 11.0 & -- & -- & -- & -- & -- & -- \\
    MedVInT~\cite{pmcvqa}$^\dagger$ {\scriptsize(Commun.\ Med.'24)}    & 44.3 & 24.6 & 40.4 & 32.6 & -- & 21.2 & -- & -- & -- & -- & -- & -- \\
    RadFM~\cite{radfm}$^\dagger$ {\scriptsize(Nat.\ Commun.'25)}       & 44.2 & 23.7 & 39.9 & 31.5 & -- & 19.5 & -- & -- & -- & -- & -- & -- \\
    CT2Rep~\cite{ct2rep}$^\dagger$ {\scriptsize(MICCAI'24)}     & 44.4 & 23.6 & 40.2 & 31.0 & -- & 13.9 & -- & -- & -- & -- & -- & -- \\
    M3D~\cite{m3d}$^\dagger$ {\scriptsize(arXiv'24)}           & 43.6 & 24.5 & 40.0 & 32.6 & -- & 14.8 & -- & -- & -- & -- & -- & -- \\
    CT-GRAPH~\cite{ctgraph}$^\S$ {\scriptsize(ICCVW'25)}       & 46.1 & 23.6 & 42.8 & 37.1 & \textbf{11.8} & 34.8 & 20.5 & 41.4 & 67.3 & 23.6 & 53.5 & 40.5 \\
    DiffVP~\cite{diffvp}$^\ddagger$ {\scriptsize(ECCV'26)}    & 48.3 & 25.2 & 45.3 & 37.6 & 5.8 & 37.9 & 20.8 & 34.4 & 67.3 & 24.2 & 54.4 & 32.9 \\
    Reg2RG~\cite{reg2rg}$^\ddagger$ {\scriptsize(TMI'25)}    & 47.3 & 24.9 & 44.2 & 38.0 & 9.1 & 23.7 & 20.9 & 37.6 & 66.9 & 14.4 & 54.4 & 64.2 \\
    \method (ours)                     & \textbf{51.1} & \textbf{29.6} & \textbf{52.4} & \textbf{41.5} & 6.8 & \textbf{59.7} & \textbf{25.6} & \textbf{49.6} & \textbf{72.5} & \textbf{42.9} & \textbf{61.6} & \textbf{20.5} \\
    \midrule
    \multicolumn{13}{@{}l}{\emph{CTRG-Chest}} \\
    Dia-LLaMA~\cite{dialama}$^\ddagger$ {\scriptsize(MICCAI'25)} & 50.3 & 30.2 & 53.1 & \textbf{45.7} & 6.4 & 41.8 & 30.6 & 41.4 & 70.8 & 24.5 & 61.8 & 37.0 \\
    Reg2RG~\cite{reg2rg} {\scriptsize(TMI'25)} & 47.3 & 25.6 & 46.8 & 31.3 & 5.1 & 28.9 & 27.7 & 34.5 & 69.1 & 18.2 & 55.5 & 50.3 \\
    \method (ours)                     & \textbf{57.5} & \textbf{37.1} & \textbf{55.1} & 36.7 & \textbf{20.9} & \textbf{46.4} & \textbf{38.4} & \textbf{44.1} & \textbf{76.9} & \textbf{33.0} & \textbf{65.4} & \textbf{27.6} \\
    \bottomrule
  \end{tabular}
\end{table*}

\section{Experiments}
\label{sec:exp}

\subsection{Experimental Settings}
\label{sec:setup}

\paragraph{Datasets.}
\textbf{RadGenome-ChestCT}~\cite{radgenome} pairs CT-RATE volumes~\cite{ctrate} with region-level reports and region masks.
We train on 23,896 training volumes, hold out another 230 training CTs for development, and evaluate on the 1,564 validation CTs used by Reg2RG~\cite{reg2rg}.
\textbf{CTRG-Chest}~\cite{ctrg} has 1,232 training, 180 validation, and 362 test chest CTs with Chinese reports and English translations.
Region masks of CTRG-Chest come from TotalSegmentator~\cite{totalseg}.
Following Reg2RG, we join the region sections of a CT in a fixed order and score the joined report against the full reference report, with dataset details in \cref{supp:data}.

\paragraph{Evaluation metrics.}
Language metrics are BLEU-1/4~\cite{bleu}, METEOR~\cite{meteor}, and ROUGE-L~\cite{rouge} under the Reg2RG protocol, plus CIDEr~\cite{cider}.
Clinical metrics are clinical efficacy F1 (CE-F1) using the CT-RATE 18-label RadBERT classifier~\cite{ctrate}, RadGraph F1~\cite{radgraph,radgraphf1}, GREEN~\cite{green}, and RaTEScore~\cite{ratescore}.
Polarity metrics structure generated and reference reports as in \cref{sec:formulation} and compare their stated observations via positive F1, negative F1, and false negations (FNeg).
Negative F1 counts a negative as correct only if the reference states it, and FNeg counts the stated negatives per 100 reports for observations present in the reference.
Scores are $\times$100 and compared by paired bootstrap over CTs~\cite{koehn2004}, with intervals in \cref{supp:eval}.

\paragraph{Baselines.}
On RadGenome-ChestCT, we take the scores of R2GenGPT~\cite{r2gengpt}, MedVInT~\cite{pmcvqa}, RadFM~\cite{radfm}, CT2Rep~\cite{ct2rep}, and M3D~\cite{m3d} from Reg2RG~\cite{reg2rg} and rescore the released predictions of Reg2RG, DiffVP~\cite{diffvp}, and CT-GRAPH~\cite{ctgraph}, whose predictions cover 1,505 CTs.
On CTRG-Chest, we rescore released Dia-LLaMA~\cite{dialama} predictions on the same test split and retrain Reg2RG with its official code.
Two further baselines isolate our design choices.
\textbf{LLaDA w/o plan} shares the \method backbone and visual prefix and writes each region section on one canvas without a plan.
\textbf{\method-AR} replaces the \method generator with an AR model (Meta-Llama-3-8B~\cite{llama3} with LoRA) that decodes greedily from the same visual prefix and the same plan given as text.
\Cref{supp:impl} details their training.

\begin{table*}[t]
  \caption{Ablation on RadGenome-ChestCT validation. (a)~Plan rule, (b1)~learned assignment, (b2)~\Cov decision, (c1)~observation canvas, (c2)~prototype, and (c3)~\Skip targets.}
  \label{tab:ablation}
  \centering
  \small
  \setlength{\tabcolsep}{3.5pt}
  \resizebox{\linewidth}{!}{%
  \begin{tabular}{@{}l c cc ccc c ccccccc@{}}
    \toprule
    & Planning & \multicolumn{2}{c}{Assignment} & \multicolumn{3}{c}{Diffusion} & & & & & & & \\
    \cmidrule(lr){2-2} \cmidrule(lr){3-4} \cmidrule(lr){5-7}
    & (a) & (b1) & (b2) & (c1) & (c2) & (c3) & Words & B-4 & CE-F1 & GREEN & RaTE & Pos-F1 & Neg-F1 & FNeg$\downarrow$ \\
    \midrule
    (1) \method & three-state & \checkmark & \checkmark & \checkmark & \checkmark & \checkmark & 247 & 29.6 & 59.7 & 49.6 & 72.5 & 42.9 & 61.6 & 20.5 \\
    (2) & two-state & \checkmark & \checkmark & \checkmark & \checkmark & \checkmark & 1108 & 8.8 & 7.1 & 17.3 & 67.1 & 17.6 & 34.2 & 52.8 \\
    (3) & three-state & \checkmark & \checkmark & \checkmark & \xmark & \checkmark & 163 & 20.1 & 19.2 & 32.5 & 61.2 & 12.1 & 48.7 & 12.9 \\
    (4) & three-state & \xmark & \checkmark & \checkmark & \checkmark & \checkmark & 221 & 29.9 & 58.7 & 50.2 & 72.9 & 43.2 & 61.0 & 23.5 \\
    (5) & three-state & \xmark & \checkmark & \checkmark & \checkmark & \xmark & 233 & 29.0 & 58.5 & 48.2 & 72.2 & 43.0 & 61.3 & 23.6 \\
    (6) & three-state & \xmark & \xmark & \checkmark & \checkmark & \checkmark & 293 & 25.1 & 59.0 & 51.1 & 73.6 & 43.5 & 61.7 & 21.2 \\
    \midrule
    (7) LLaDA w/o plan & -- & \xmark & \xmark & \xmark & \xmark & \xmark & 231 & 27.4 & 37.7 & 37.6 & 69.6 & 28.8 & 60.0 & 30.0 \\
    \bottomrule
  \end{tabular}}
\end{table*}

\paragraph{Implementation details.}
Observation vocabularies are built with locally run Qwen3.5-9B~\cite{qwen35} and Qwen3.8-27B~\cite{qwen38}.
At inference, ridge regression predicts segment lengths from the region feature and planned state.
The three-state classifiers are cross-fitted in 5 folds grouped by patient.
Positive thresholds require out-of-fold precision of at least 0.3.
We set $\kappa = 0.3$ and $\lambda = 4$ on all datasets, so a planned negative has a positive share below 20\%.
On RadGenome-ChestCT, each $\kappa \in \{0.3, 0.4, 0.5\}$ outperforms all prior methods on every metric except CIDEr.
The frozen NV-Reason-CT encoder was trained on CT-RATE training volumes, never on CTRG-Chest.
The generator is LLaDA-8B-Instruct~\cite{llada} with rank-8 LoRA, trained on RadGenome-ChestCT for 3 epochs with $f = 0.3$ and E-steps before epochs 2 and 3.
\Cref{supp:impl,supp:results} give the $\kappa$ check, decoding settings, hardware, and costs.

\subsection{Comparison with State-of-the-Art Methods}
\label{sec:sota}

On RadGenome-ChestCT, \method leads every metric in \cref{tab:sota} except CIDEr, with its largest gains on metrics that score findings.
Over Reg2RG, it raises positive F1 from 14.4 to 42.9 and cuts false negations from 64.2 to 20.5 per 100 reports.
Reg2RG recovers only 11\% of reference findings and either leaves the rest unmentioned or denies them.
In \method, the plan instead states a finding only when its classifier supports it and a negative only when the expected-cost rule accepts the risk.
Without this rule, CT-GRAPH and DiffVP still make 1.6 to 2 times as many false negations.
\Cref{supp:results} also reports \method with other visual encoders.
CIDEr is lower because its length penalty dominates the score of a full report and \method's reports are on average 53 words longer than the references.
Without this penalty, \method scores highest, as \cref{supp:eval} shows.
On CTRG-Chest, \method leads every metric except ROUGE-L and raises GREEN from 41.4 for Dia-LLaMA to 44.1.
Dia-LLaMA keeps the reference sentence order, and grouping its sentences by region lowers its ROUGE-L to 31.5.

\begin{table*}[t]
  \caption{\method-AR and \method under the same plans, visual prefix, and training examples.}
  \label{tab:mdlm_ar}
  \centering
  \small
  \setlength{\tabcolsep}{4.5pt}
  \begin{tabular}{@{}ll cccc cccc cc@{}}
    \toprule
    Dataset & Generator & B-4 & R-L & CIDEr & CE-F1 & RadGraph & GREEN & RaTE & Pos-F1 & Neg-F1 & FNeg$\downarrow$ \\
    \midrule
    \multirow{2}{*}{RadGenome-ChestCT} & \method-AR & 28.0 & 40.3 & \textbf{8.6} & 57.2 & 23.8 & 48.3 & 71.4 & 42.3 & \textbf{61.6} & \textbf{19.4} \\
     & \method & \textbf{29.6} & \textbf{41.5} & 6.8 & \textbf{59.7} & \textbf{25.6} & \textbf{49.6} & \textbf{72.5} & \textbf{42.9} & \textbf{61.6} & 20.5 \\
    \midrule
    \multirow{2}{*}{CTRG-Chest} & \method-AR & 33.5 & 35.4 & 12.3 & \textbf{48.9} & 35.5 & 42.1 & 74.0 & \textbf{33.5} & 65.0 & 29.6 \\
     & \method & \textbf{37.1} & \textbf{36.7} & \textbf{20.9} & 46.4 & \textbf{38.4} & \textbf{44.1} & \textbf{76.9} & 33.0 & \textbf{65.4} & \textbf{27.6} \\
    \bottomrule
  \end{tabular}
\end{table*}

\subsection{Ablation Study}
\label{sec:ablation}

\begin{figure}[t]
  \centering
  \includegraphics[width=0.9\linewidth]{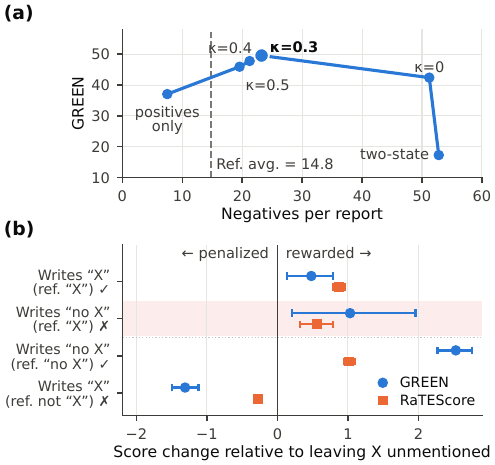}
  \caption{(a)~Negatives per report and GREEN as $\kappa$ varies, with the dashed line at the reference mean. ``Positives only'' still states negatives because a region without a planned positive falls back to its most likely observation. (b)~Score change for each way of stating X, relative to leaving X unmentioned, with rows labeled by what the reference states.}
  \label{fig:analysis}
\end{figure}

\paragraph{Effect of three-state constraint planning.}
In (2) of \cref{tab:ablation}, the two-state rule plans a negative for every observation whose positive probability is below 0.2, so many findings missed by the classifier are denied, and false negations rise from 20.5 to 52.8 per 100 reports.
It states 52.8 negatives per report, which makes reports 4.5 times longer while GREEN falls from 49.6 to 17.3.
In \cref{fig:analysis}a, lowering $\kappa$ from 0.5 to 0.3 adds negatives while negative F1 barely changes, and GREEN rises from 45.9 to 49.6 because it credits correct negatives, as in \cref{fig:analysis}b.
At $\kappa = 0$, the plan states 51.2 negatives per report, more than three times the 14.8 in the references.
Negative F1 then falls to 37.8 and GREEN to 42.4, so extra negatives help only while they stay accurate.
At an equal number of planned negatives, a single threshold on $p_{i1}$ plans 20.3 false negations per 100 reports, whereas the expected-cost rule plans 12.0, and \cref{supp:results} compares these rules before and after generation.

\paragraph{Effect of latent sentence assignment.}
Over the first-observation assignment in (4), the learned assignment raises CE-F1 from 58.7 to 59.7 and cuts false negations from 23.5 to 20.5.
This gain comes from recall, and lymphadenopathy explains most of the extra true positives.
When mediastinal lymphadenopathy (LAP) is planned as present, the first-observation model writes only the frequent negative ``No pathological LAP was detected in the mediastinum'' in 39\% of the LAP segments, whereas the learned assignment does so in 12\%.
Without the \Cov decision in (6), every segment must write its observation.
The model thus repeats multi-observation sentences and lengthens reports from 221 to 293 words, which GREEN does not penalize.

\paragraph{Effect of position-constrained canvas diffusion.}
LLaDA w/o plan in (7) of \cref{tab:ablation} lets the generator choose findings on its own, and CE-F1 falls from 59.7 to 37.7.
The same plans given as a text prompt on one canvas of equal length reach a CE-F1 of only 47.8.
Thus the plan and observation canvas, rather than the diffusion backbone, drive most of the gain over a plain MDLM.
The generator states the planned state in 92\% of the written positive segments, and \cref{supp:results} examines its deviations and its use of the visual prefix.
Because the plan reaches the generator only via prototypes (\cref{eq:position}), removing them at inference in (3) strips each segment of its observation and state.
The generator then misses 1.4 more findings per report, and positive F1 falls from 42.9 to 12.1.
\Skip segments train the generator on planned observations that the reference does not write.
Without them in (5), the share of empty segments stays at about 30\%, but each written segment becomes longer, so reports grow from 221 to 233 words.
GREEN then drops significantly by 2.0 from (4), as it counts 0.27 more false findings per report.

\subsection{Analysis}
\label{sec:analysis}

\begin{figure}[t]
  \centering
  \includegraphics[width=0.85\linewidth]{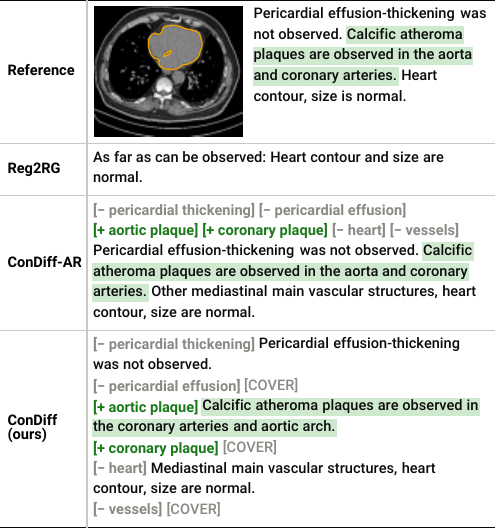}
  \caption{A heart section. Highlights link the same finding, red marks wrong statements, and \covtok marks a covered segment. Chips show each planned observation and state, which \method-AR receives together as a text prompt.}
  \label{fig:cases}
\end{figure}

\paragraph{Masked diffusion as the generator.}
\label{sec:mdlm_ar}
\Cref{tab:mdlm_ar} compares the generators under the same visual prefix, training examples, and plans.
On both datasets, \method achieves significantly higher GREEN and RaTEScore, mainly because it reduces the clinically significant false findings in GREEN's counts from 2.43 to 2.25 per report on RadGenome-ChestCT and from 2.07 to 1.73 on CTRG-Chest.
We attribute this reduction to the observation canvas, which also raises GREEN for an AR model, as \cref{supp:results} shows.
On CTRG-Chest, \method-AR has higher CE-F1 and positive F1, partly because \method learns its assignment from only 1,232 CTs, as \cref{supp:results} shows.

\paragraph{False negations in existing metrics.}
In \cref{fig:analysis}b, we regress GREEN and RaTEScore on statement counts per type across 7,820 validation reports from five systems, with one intercept per CT and one per system.
Each coefficient gives the change from stating observation X in a given way relative to leaving X unmentioned on the same CT.
A correct negative comes with a GREEN higher by 2.5 and a false positive with one lower by 1.3.
Yet a false negation comes with a GREEN higher by 1.0, as much as stating X correctly, and a RaTEScore higher by 0.6.
CE-F1 cannot separate a false negation from leaving X unmentioned either, as its binary labels treat both as negative.
On generated reports, denying a present finding scores higher than leaving it unmentioned, so we report false negations separately.

\paragraph{Qualitative results.}
\label{sec:qualitative}
\Cref{fig:cases} shows a heart section whose reference states calcific plaques in the aorta and coronary arteries and no pericardial effusion or thickening.
Reg2RG states only that the heart is normal and misses both plaques, whereas \method-AR and \method receive the same plan and state every planned observation correctly.
In \method, three sentences each state two planned observations, so the segments for pericardial effusion, coronary plaque, and vessels start with \Cov instead of repeating them.
\Cref{supp:examples} gives more examples and a failure case.

%% file: sec/5_conclusion.tex
\section{Conclusion}
\label{sec:conclusion}
We presented \method, to our knowledge the first MDLM for 3D CT report generation, which sets a three-state constraint for each observation and gives each planned observation its own canvas segment.
It makes fewer false negations and misses fewer positive findings than prior methods, and significantly outperforms an AR model on GREEN and RaTEScore under the same plans.

%% file: sec/X_suppl.tex
\twocolumn[{\centering\Large\bf Appendix\par\vspace{1.5em}}]
\appendix
\raggedbottom  %

\section{Datasets and Preprocessing}
\label{supp:data}

\paragraph{RadGenome-ChestCT.}
We train on 23,896 training volumes, hold out another 230 training CTs for development, and evaluate on the 1,564 validation CTs used by Reg2RG.
Reference reports state 14.8 negatives and 6.0 positives per report, so negatives account for 71\% of the observation statements.
Of the 379,470 training sentences, 42.2\% state two or more observations, such as ``no pericardial effusion or thickening''.

\paragraph{CTRG-Chest.}
The volumes are released as 8-bit slices, which we map to approximate HU by $4.706\,g - 1000$ for a gray value $g$, clipping values above 200 HU.
We correct the orientation following CT-CLIP~\cite{ctrate} and estimate the voxel spacing from the lung size as 0.83, 0.83, and 1.03\,mm, because the 1.5\,mm slice thickness in the headers is incorrect.
We drop the breast and thyroid, which lack masks, leaving 8 regions.
CTRG-Chest has no effusion masks, so the $N_r$ of \cref{supp:method} uses the region masks alone.
DeepSeek-V4-Flash assigns each English report sentence to a region, and we remove head CT paragraphs and image references mixed into the reports.
From the released split of 1,262, 180, and 362 CTs, we drop 29 training CTs from patients who also appear in validation or testing and 1 training CT without region text, leaving 1,232 training CTs.
We do not compare with the CTRG-Chest numbers reported by Reg2RG because its split places 250 of its 361 test CTs in our training set.

\Cref{tab:supp_data} summarizes the two datasets.

\begin{table}[H]
  \caption{Dataset statistics. Counts per report are on each test set, which for RadGenome-ChestCT is its validation split.}
  \label{tab:supp_data}
  \centering
  \footnotesize
  \setlength{\tabcolsep}{2.5pt}
  \begin{tabular}{@{}lcc@{}}
    \toprule
    & RadGenome & CTRG \\
    \midrule
    Training CTs & 23,896 & 1,232 \\
    Development or validation CTs & 230 & 180 \\
    Test CTs & 1,564 & 362 \\
    Observations & 192 & 239 \\
    Training sentences & 379,470 & 8,544 \\
    Positives per report & 6.0 & 4.7 \\
    Negatives per report & 14.8 & 10.4 \\
    \bottomrule
  \end{tabular}
\end{table}

\section{Method Details}
\label{supp:method}

\paragraph{Sentence structuring.}
Report sentences are first normalized into phrasings by lowercasing, replacing numbers with a placeholder, and removing lead-in clauses, yielding 105,907 distinct phrasings in RadGenome-ChestCT.
Using the prompt in \cref{fig:supp_prompt}, Qwen3.5-9B~\cite{qwen35} structures each phrasing from its most frequent original wording together with the sentence that most often precedes it to resolve references such as ``it''.
Each statement becomes an (anatomy, finding, status) triple, where the status is present, absent, normal, uncertain, or not evaluable, together with a type such as observation, impression, or comparison.

\paragraph{Observation vocabulary.}
Qwen3.8-27B~\cite{qwen38} constructs the vocabulary in three steps, each restricted to a list of valid answers.
It maps each of the 775 anatomy names to one of 39 organs, proposes finding categories for each organ from that organ's frequent finding names, and maps each of the 5,981 (organ, finding) names to one category or to ``other''.
An observation is an (organ, category) pair, and vessel wall atherosclerosis, calcification, and plaque form a single category because reports use these words interchangeably.
Only statements of the observation type and technique statements that mark a structure as not evaluable are counted, so impressions, recommendations, comparisons with prior examinations, and other technique statements are excluded.
Present and uncertain statements map to the positive state and absent statements map to the negative state.
Normal and not evaluable statements form their own observations, such as (heart, normal) and (heart, not evaluable), which, like the normal statements in \cref{sec:formulation}, never take state 2.
For example, ``Bilateral adrenal glands were normal and no space-occupying lesion was detected.'' states (adrenal glands, normal) and (adrenal glands, mass), both negative.
Observations stated in fewer than 1\% of the training reports of a region are folded into an ``other'' observation for their organ or region, leaving 192 observations, of which 158 are findings.
The same steps give 239 observations on CTRG-Chest, of which 195 are findings, although it has far fewer training sentences.
Because the 1\% threshold is relative to the reports of each region, and the esophagus and the trachea and bronchi have sections in only 39 and 162 CTRG-Chest training reports, the threshold keeps observations stated once or twice there.
These sections also mention other organs, so the two regions hold 82 of the 239 observations.
When a region section states an observation as both negative and positive, the observation takes the positive state, as in \cref{eq:state}.

\begin{figure*}[!t]
  \centering
  \fbox{\parbox{0.97\textwidth}{\scriptsize\ttfamily\raggedright
  \textnormal{\textbf{System.}} You structure sentences of chest CT radiology reports (CT-RATE). Each sentence comes with the report region it belongs to. Decompose the sentence into atomic observations: one per (structure, finding) it states. The sentence may come with the preceding sentence of the same report: use it only to resolve references ("it", "this lesion", "the described findings"); never structure the preceding sentence itself. Return only JSON: \{"observations": [\{"anatomy": ..., "finding": ..., "status": ..., "location": ..., "size": ..., "attributes": [...], "type": ...\}]\}
\par\smallskip
- anatomy: the organ or structure observed, short lowercase noun phrase without laterality, lobe or segment (e.g. "lung", "pleura", "heart", "pericardium", "coronary arteries", "thoracic aorta", "mediastinal lymph nodes", "trachea", "bronchi", "esophagus", "thyroid", "liver", "gallbladder", "kidney", "vertebrae", "ribs", "breast").
\newline - finding: what is observed, a short canonical radiology term, lowercase, singular, without location, size or severity (e.g. "emphysema", "nodule", "ground-glass opacity", "consolidation", "atelectasis", "interlobular septal thickening", "bronchiectasis", "fibrotic sequela", "calcification", "effusion", "enlargement", "hiatal hernia", "degenerative change", "lymphadenopathy"). Use "normal" when the structure is only described as normal, natural or unremarkable.
\newline - status: "present" (stated), "absent" (explicitly denied, e.g. "no effusion"), "normal" (finding = "normal"), "uncertain" (suspected, possible, cannot be excluded), or "not\_evaluable" (the structure is outside the field of view or cannot be assessed, e.g. without contrast; finding = "not evaluable"). status is exactly one of these five words: recommendations, comparisons ("stable", "regressed") and impressions are expressed by type, with status "present" (or "uncertain" when hedged).
\newline - location: laterality / lobe / segment / zone as written, or "".
\newline - size: measurements as written (e.g. "4 mm", "\# mm" when masked), or "".
\newline - attributes: other descriptors (e.g. "mild", "linear", "calcified", "diffuse", "peripheral"), or [].
\newline - type: "observation" (something seen on this CT), "impression" (a diagnostic interpretation, e.g. "in favor of covid-19 pneumonia", "nonspecific"), "recommendation" (follow-up, correlation), "comparison" (change from a prior exam, e.g. "stable"), or "technique" (exam limitation, e.g. "evaluated suboptimally without contrast").
\newline A sentence denying several findings gives one "absent" observation per finding. Keep the report's meaning; do not add findings; if the sentence states nothing about a structure, return \{"observations": []\}.
  \par\smallskip
  \textnormal{\textbf{User.}} Region: heart\newline Sentence: Heart contour and size are normal. No pericardial effusion-thickening was detected.
  \par\smallskip
  \textnormal{\textbf{Assistant.}} \{"observations": [\{"anatomy": "heart", "finding": "normal", "status": "normal", "location": "", "size": "", "attributes": [], "type": "observation"\}, \{"anatomy": "pericardium", "finding": "effusion", "status": "absent", "location": "", "size": "", "attributes": [], "type": "observation"\}, \{"anatomy": "pericardium", "finding": "thickening", "status": "absent", "location": "", "size": "", "attributes": [], "type": "observation"\}]\}
  }}
  \caption{Prompt that structures a report sentence for Qwen3.5-9B, followed by one of its six worked examples.}
  \label{fig:supp_prompt}
\end{figure*}

\paragraph{Label check.}
CT-RATE assigns 18 abnormality labels to every CT, which are predicted from the same reports by RadBERT rather than annotated by radiologists.
After mapping each label to its observations, our structured labels agree with the CT-RATE labels on the 1,564 validation CTs with a micro F1 of 89.2 and a macro Cohen's kappa of 85.8, as shown in \cref{tab:supp_label_check}.
Of the 5,358 positive (CT, label) pairs in CT-RATE, our structure states 1.8\% as negative, mostly for lymphadenopathy.
In 67 of those 70 lymphadenopathy reports, the sentence denies pathologically enlarged lymph nodes, as in ``No pathological LAP was detected in the mediastinum'', while the other 3 compare lymph nodes with a prior examination.
One of these 3, ``There was no significant difference in metastatic lymph node dimensions in the left supraclavicular fossa'', reports metastatic lymph nodes that our structure wrongly states as negative.
A further 12.3\% of the positive pairs have no matching statement in our structure, 45\% of which occur for pulmonary fibrotic sequela, lung opacity, and arterial wall calcification.

\begin{table*}[!t]
  \caption{Agreement of the structured labels with the 18 CT-RATE labels on the 1,564 validation CTs. Positives are counted per CT. Neg.\ is the number of CT-RATE positives that our structure states as negative.}
  \label{tab:supp_label_check}
  \centering
  \small
  \setlength{\tabcolsep}{6pt}
  \begin{tabular}{@{}lccccccc@{}}
    \toprule
    Label & CT-RATE pos. & Ours pos. & Prec. & Rec. & F1 & $\kappa$ & Neg. \\
    \midrule
    Medical material & 158 & 148 & 83.8 & 78.5 & 81.0 & 79.0 & 0 \\
    Arterial wall calcification & 435 & 394 & 97.2 & 88.0 & 92.4 & 89.7 & 0 \\
    Cardiomegaly & 164 & 154 & 92.2 & 86.6 & 89.3 & 88.1 & 0 \\
    Pericardial effusion & 110 & 117 & 82.9 & 88.2 & 85.5 & 84.3 & 2 \\
    Coronary artery wall calcification & 388 & 352 & 97.7 & 88.7 & 93.0 & 90.8 & 0 \\
    Hiatal hernia & 218 & 201 & 99.0 & 91.3 & 95.0 & 94.2 & 0 \\
    Lymphadenopathy & 398 & 294 & 99.3 & 73.4 & 84.4 & 80.1 & 70 \\
    Emphysema & 305 & 311 & 96.5 & 98.4 & 97.4 & 96.8 & 0 \\
    Atelectasis & 364 & 422 & 82.5 & 95.6 & 88.5 & 84.7 & 0 \\
    Lung nodule & 704 & 657 & 97.3 & 90.8 & 93.9 & 89.2 & 14 \\
    Lung opacity & 606 & 518 & 98.1 & 83.8 & 90.4 & 85.1 & 1 \\
    Pulmonary fibrotic sequela & 427 & 305 & 91.1 & 65.1 & 76.0 & 68.9 & 0 \\
    Pleural effusion & 189 & 158 & 96.2 & 80.4 & 87.6 & 86.1 & 6 \\
    Mosaic attenuation pattern & 126 & 125 & 96.0 & 95.2 & 95.6 & 95.2 & 1 \\
    Peribronchial thickening & 175 & 155 & 88.4 & 78.3 & 83.0 & 81.0 & 0 \\
    Consolidation & 298 & 293 & 91.5 & 89.9 & 90.7 & 88.5 & 1 \\
    Bronchiectasis & 169 & 201 & 80.6 & 95.9 & 87.6 & 85.9 & 0 \\
    Interlobular septal thickening & 124 & 158 & 70.9 & 90.3 & 79.4 & 77.4 & 0 \\
    \bottomrule
  \end{tabular}
\end{table*}

\paragraph{Region masks.}
RadGenome-ChestCT provides masks of its ten regions for every CT.
Its pleura mask is identical to its lung mask and covers only the aerated lung, so pleural effusion lies outside both.
We therefore join the pleural effusion mask to the pleura mask and the pericardial effusion mask to the heart mask, taking both from the TotalSegmentator~\cite{totalseg} segmentations released with CT-RATE~\cite{ctrate}.

\paragraph{Encoder grid.}
For the encoder input, we follow the NV-Reason-CT loader, which reorients each volume to LPS, resamples it to 2\,mm, clips intensities to $[-1000, 1000]$\,HU, and crops $192^3$ voxels.
The crop is centered in plane and, along the craniocaudal axis, centered on the lung extent detected from enclosed air.
The lungs are detected in 1,552 of the 1,564 validation CTs, whereas the crop for the remaining 12 starts at the superior edge.
With $8^3$ patches, the encoder tokens lie on a $24^3$ grid of 16\,mm cells.

\paragraph{Region cells.}
Each cropped voxel is assigned the mask label of its nearest voxel in the RadGenome-ChestCT volume, and $\omega_c$ denotes the fraction of the 512 voxels in cell $c$ that lie inside the region mask.
$N_r$ contains the cells touched by the region mask joined with its effusion mask, dilated by one cell.
\Cref{tab:supp_masks} shows that the thyroid, esophagus, and trachea touch 20 to 61 cells on average yet cover only 1.1 to 4.2 of them by at least half, so this dilation allows their features to include the surrounding tissue.
The effusion masks add cells to the pleura in 18.7\% of the validation CTs and to the heart in 52.1\%.

\begin{table}[t]
  \caption{Region masks on the $24^3$ encoder grid in the 1,564 validation CTs of RadGenome-ChestCT. No mask is the share of CTs without the mask, and cell counts are averaged over the CTs with the mask. A touched cell contains any mask voxel, and a half-covered cell has $\omega_c \ge 0.5$.}
  \label{tab:supp_masks}
  \centering
  \footnotesize
  \setlength{\tabcolsep}{2.5pt}
  \begin{tabular}{@{}lcccc@{}}
    \toprule
    Region & No mask (\%) & Touched & Half-covered & $\lvert N_r \rvert$ \\
    \midrule
    Abdomen & 0.1 & 1,620 & 933 & 3,441 \\
    Bone & 0.0 & 1,595 & 168 & 4,629 \\
    Breast & 12.6 & 134 & 47 & 470 \\
    Esophagus & 0.6 & 55 & 2.8 & 307 \\
    Heart & 0.7 & 295 & 144 & 820 \\
    Lung & 0.0 & 1,729 & 1,027 & 3,641 \\
    Mediastinum & 0.4 & 435 & 84 & 1,500 \\
    Pleura & 0.0 & 1,729 & 1,027 & 3,677 \\
    Thyroid & 2.2 & 20 & 1.1 & 115 \\
    Trachea and bronchi & 0.3 & 61 & 4.2 & 351 \\
    \bottomrule
  \end{tabular}
\end{table}

\paragraph{Region feature and three-state classifiers.}
The region feature concatenates three means of the frozen encoder tokens $\mathbf{e}_c$ on the $24^3$ grid,
\begin{equation}
  \begin{split}
  \mathbf{h}_r = \Big[&\tfrac{\sum_{c \in N_r} \omega_c \mathbf{e}_c}{\sum_{c \in N_r} \omega_c};\ \tfrac{1}{|N_r|} \textstyle\sum_{c \in N_r} \mathbf{e}_c; \\
  &\tfrac{1}{24^3} \textstyle\sum_{c} \mathbf{e}_c\Big] \in \mathbb{R}^{2592},
  \end{split}
\end{equation}
where $\mathbf{e}_c$ is layer normalized and $N_r$ and $\omega_c$ are defined above.
We standardize $\mathbf{h}_r$ within each training fold to obtain $\tilde{\mathbf{h}}_r$, and the classifier for observation $i$ is
\begin{equation}
  \begin{split}
  &\mathbf{p}_i = \operatorname{softmax}\big(W_i^\top \tilde{\mathbf{h}}_r + \mathbf{b}_i\big), \\
  &\min_{W_i, \mathbf{b}_i} \frac{1}{|\mathcal{V}|} \sum_{v \in \mathcal{V}} \mathrm{CE}\big(y_i^{(v)}, \mathbf{p}_i^{(v)}\big) + \frac{\lVert W_i \rVert_F^2}{2C|\mathcal{V}|},
  \end{split}
\end{equation}
where $\mathcal{V}$ is the set of training CTs, $y_i^{(v)}$ is the state of observation $i$ in CT $v$ from \cref{eq:state}, $C = 0.002$, and the bias is unpenalized.
The region gate is $g_r(V) = \mathds{1}\big[\sigma(\mathbf{w}_r^\top \tilde{\mathbf{h}}_r + b_r) \ge \tau_r\big]$ with $\tau_r$ set at the out-of-fold F1 optimum.
On the observations that the classifier can state, $\hat{r} = p_{i2} / (p_{i1} + p_{i2})$ is well calibrated, with an expected calibration error of 0.0011 out of fold and 0.0037 on the development CTs.
The positive threshold is
\begin{equation}
  \eta_i = \arg\max_\eta \mathrm{F1}^{\text{oof}}_i(\eta) \quad \text{s.t.} \quad \mathrm{Prec}^{\text{oof}}_i(\eta) \ge 0.3.
\end{equation}
Observations with fewer than 20 training positives share a pooled threshold, and observations that do not reach a precision of 0.3 at any threshold are never planned as positive.
On RadGenome-ChestCT, this applies to 54 of the 158 finding observations, which account for 7.4\% of the positive statements in the training reports.

\paragraph{Plan assembly.}
The canonical position of observation $i$ is the mean relative position of its sentences across the training region sections,
\begin{equation}
  \rho(i) = \operatorname{mean}\Big\{\tfrac{k-1}{\max(|S| - 1,\, 1)} : u_k \in S,\ i \in C(u_k),\ S \in \mathcal{D}_{\text{train}}\Big\},
\end{equation}
where $u_k$ is the $k$-th of the $|S|$ sentences in region section $S$, so the sentence of a one-sentence section is at position 0.
The plan of a region lists the selected observations sorted by $\rho$,
\begin{equation}
  \begin{split}
  P_r = \operatorname{sort}_\rho\Big(&\{(i, 2) : p_{i2} > \eta_i\} \cup \{(i, 1) : p_{i2} \le \eta_i, \\
  &p_{i1} > \lambda p_{i2} + \kappa p_{i0},\ i \in \mathcal{I}_r^-\}\Big),
  \end{split}
\end{equation}
where $\mathcal{I}_r^-$ contains the observations that at least one training report states as negative.
If a region passes the gate with an empty plan, its plan instead holds the single observation $i^\star = \arg\max_i (1 - p_{i0})$ in its more likely stated state.
\Cref{tab:supp_cost} gives the costs behind \cref{eq:rule}, where $m$ is the cost of a missed finding.
With $\kappa = -1$, the condition $p_{i1} > \lambda p_{i2} + \kappa p_{i0}$ reduces to $p_{i2} < 1/(1 + \lambda)$, which is the two-state rule in \cref{tab:ablation} row (2).

\begin{table}[H]
  \caption{Costs of stating ``no $i$'' or leaving $i$ unmentioned, by the state of observation $i$ in the reference.}
  \label{tab:supp_cost}
  \centering
  \small
  \begin{tabular}{@{}lccc@{}}
    \toprule
    Action & Unmentioned & Negative & Positive \\
    \midrule
    State ``no $i$'' & $\kappa$ & $0$ & $\lambda + m$ \\
    Leave $i$ unmentioned & $0$ & $1$ & $m$ \\
    \bottomrule
  \end{tabular}
\end{table}

\paragraph{Segment lengths.}
During training, the segment length for observation $i$ is
\begin{equation}
  n_i = 20 \cdot \min\Big(5,\ \max\big(1, \lceil (Q_i + 1)/20 \rceil\big)\Big),
\end{equation}
where $Q_i$ is the 95th percentile of the token lengths of its training texts, and longer texts are truncated to $n_i - 1$ tokens.
At inference, a ridge regression for each observation predicts the log of the text length plus one \eos token from the region feature and the planned state, and we exponentiate the prediction plus the 95th percentile of its residuals before rounding up in the same way.
The ridge penalty and the residual percentile are calibrated on a held-out 20\% of the training patients.
A training canvas holds at most 500 tokens, beyond which the least likely observations are dropped.
\Cref{tab:supp_seglen} shows that the image contributes little to segment length.
On the same observations, the $R^2$ of the log length is 0.538 from the observation and state alone and 0.559 after adding the image on RadGenome-ChestCT, and 0.599 and 0.602 on CTRG-Chest.

\begin{table}[H]
  \caption{Truncated segments and mean segment length in tokens on held-out CTs. The first two rows use the 95th percentile of training lengths per observation, or per observation and state. True length rounds the reference length up and is a lower bound.}
  \label{tab:supp_seglen}
  \centering
  \small
  \setlength{\tabcolsep}{4pt}
  \begin{tabular}{@{}lcccc@{}}
    \toprule
    & \multicolumn{2}{c}{RadGenome} & \multicolumn{2}{c}{CTRG} \\
    \cmidrule(lr){2-3} \cmidrule(l){4-5}
    Segment length & Trunc. & Length & Trunc. & Length \\
    \midrule
    Per observation & 1.9\% & 42.1 & 3.5\% & 56.9 \\
    Per observation and state & 1.6\% & 39.6 & 4.0\% & 55.1 \\
    Ridge regression & 1.5\% & 40.8 & 3.7\% & 56.8 \\
    True length & 0 & 27.0 & 0 & 37.9 \\
    \bottomrule
  \end{tabular}
\end{table}

\paragraph{Training canvases and corruption.}
A training canvas contains the observations from the reference region section together with likely observations that the reference leaves unmentioned,
\begin{equation}
  \begin{split}
  \tilde{P}_r = \operatorname{sort}_\rho\Big(&\{(i, \tilde{y}_i) : y_i > 0\} \\
  &\cup \{(i, \hat{s}_i) : y_i = 0,\ 1 - p^{\text{oof}}_{i0} \ge 0.25\}\Big),
  \end{split}
\end{equation}
where $\tilde{y}_i = 3 - y_i$ with probability 0.1 and $y_i$ otherwise, and $\hat{s}_i = 1 + \mathds{1}[p_{i2} > p_{i1}]$.
The second set defines the \Skip segments.
With $b \sim \mathrm{Bern}(f)$ and $t \sim \mathcal{U}(0, 1)$, position $j$ is masked with probability $1$ if $b = 1$ or $x_0^j = \text{\eos}$, and with probability $t$ otherwise.
The loss weights are 1 for text, 2 for \eos, and 0.25 for \void positions.

\paragraph{Hard EM.}
The expected loss of an assignment and its approximation are
\begin{align}
  L(z) &= f\,\ell_1(z) + (1 - f)\,\mathbb{E}_{t \sim \mathcal{U}(0,1)}\,\ell_t(z), \\
  \hat{L}(z) &= f\,\ell_1(z) + (1 - f)\,\ell_{1/2}(z; \boldsymbol{\xi}),
\end{align}
where $\ell_1$ is the loss on the fully masked canvas and $\ell_{1/2}$ masks the positions $\{j : \xi_j < 1/2\}$ along with every \eos using a single draw of $\boldsymbol{\xi}$, with one $\xi_j \sim \mathcal{U}(0, 1)$ per canvas position, for each region section and shared across all candidates.
The E-step computes $\ell_1$ for all candidates in one forward pass and evaluates $\ell_{1/2}$ for the two best candidates under $\ell_1$ as well as the first-observation assignment $z^{\text{first}}_u = \arg\min_{i \in C'(u)} \rho(i)$ of \cref{sec:assignment}, so it needs at most four forward passes per region section.
A region section has a median of 2 candidate assignments, a 99th percentile of 12, and a maximum of 216.
In 26.6\% of the training region sections, no candidate assignment preserves sentence order.
\Cref{tab:supp_estep} reports how many training sentences differ from the first-observation assignment after each E-step.

\begin{table}[H]
  \caption{Training sentences whose segment differs from the first-observation assignment after each E-step.}
  \label{tab:supp_estep}
  \centering
  \footnotesize
  \setlength{\tabcolsep}{2.5pt}
  \begin{tabular}{@{}lccc@{}}
    \toprule
    Dataset & First E-step & Second E-step & Sentences \\
    \midrule
    RadGenome-ChestCT & 9,175 (2.4\%) & 11,767 (3.1\%) & 379,470 \\
    CTRG-Chest & 621 (7.3\%) & 1,054 (12.3\%) & 8,544 \\
    \bottomrule
  \end{tabular}
\end{table}

\paragraph{Visual prefix and \void adapter.}
NV-Reason-CT outputs the 864-dimensional tokens $\mathbf{e}_c$ on the $24^3$ grid, from which the visual prefix is constructed as
\begin{align}
  \mathbf{G} &= \mathrm{PR}\big(\mathrm{Pool}_2(\{\mathrm{LN}(\mathbf{e}_c) + \mathrm{PE}_c\}_c)\big), \\
  \mathbf{R}_r &= \mathrm{PR}\big(\{\mathrm{LN}(\mathbf{e}_c) + \mathrm{PE}_c + \boldsymbol{\gamma}_c^\top U : c \in N_r\}\big), \\
  \mathbf{c}_r &= \big[\mathrm{LN}(W_v \mathbf{G});\ \mathrm{LN}(W_v \mathbf{R}_r);\ \mathrm{LN}(W_m \mathbf{d}_r); \nonumber \\
  &\qquad \mathbf{0};\ \mathrm{Emb}(\text{name}_r)\big],
\end{align}
where PR is the shared Perceiver resampler with 6 layers and 32 latents, LN is layer normalization, PE is a fixed 3D sinusoidal position code, $\mathrm{Pool}_2$ pools the grid to $12^3$, $\boldsymbol{\gamma}_c \in \mathbb{R}^2$ holds the mask coverage of cell $c$, $U$, $W_v$, and $W_m$ are learned projections, $\mathbf{d}_r$ is the 255-dimensional mask vector from Reg2RG, $\mathbf{0}$ is a zero token, and Emb embeds the region name.
The \void logit adds a learned correction to the frozen output layer, $\mathbf{o}_{\text{\void}}^\top \mathbf{v} + \boldsymbol{\delta}^\top \mathbf{v} + \beta$ for a final hidden state $\mathbf{v}$, where only $\boldsymbol{\delta}$ and $\beta$ are trained.

\paragraph{Prototypes.}
The set $T_{i,s}$ in \cref{eq:proto} contains the 500 most frequent phrasings, or all of them if there are fewer, and any observation and state pair without training text falls back to the prototype of the opposite state or to the zero vector.

\paragraph{Decoding within a segment.}
At each forward pass, decoding sets the probabilities of \mask and \covtok to zero at all masked positions of the current block.
It also sets the probability of \eos to zero unless every earlier position in the segment is filled and the segment already contains text, and it sets the probability of \void to zero until the segment contains an \eos.
Decoding commits every position whose top probability reaches 0.9, or the single most confident position if none reaches 0.9, and fills all positions after an \eos with \void.

\section{Implementation Details}
\label{supp:impl}

\paragraph{Hyperparameters.}
\Cref{tab:supp_hparams} lists all hyperparameter settings.

\begin{table}[H]
  \caption{Hyperparameters of \method.}
  \label{tab:supp_hparams}
  \centering
  \footnotesize
  \setlength{\tabcolsep}{2.5pt}
  \begin{tabular}{@{}ll@{}}
    \toprule
    Setting & Value \\
    \midrule
    Classifiers & $C = 0.002$, 5 folds by patient, precision floor 0.3 \\
    Plan & $\lambda = 4$, $\kappa = 0.3$ \\
    Visual prefix & 6 layers, 32 latents, 32 global and 32 region tokens \\
    LoRA & rank 8, $\alpha = 32$, dropout 0.05, \\
    & on all attention and FFN projections \\
    Training canvas & state flips 0.1, \Skip if $1 - p_{i0} \ge 0.25$ \\
    Corruption & $f = 0.3$, weights 1, 2, 0.25 for text, \eos, \void \\
    Prototypes & 500 most frequent phrasings \\
    Hard EM & E-steps before epochs 2 and 3, top 2 under $\ell_1$ \\
    Optimizer & AdamW, lr $5 \times 10^{-5}$ LoRA, $10^{-4}$ others \\
    & weight decay 0.01, 100 warmup steps, batch 24 \\
    Epochs & 3 on RadGenome-ChestCT, 6 on CTRG-Chest \\
    Decoding & blocks of 20 tokens, threshold 0.9 \\
    Hardware & one NVIDIA RTX PRO 6000 with 96\,GB \\
    \bottomrule
  \end{tabular}
\end{table}

\begin{table*}[!t]
  \caption{Visual encoders, language decoders, and extra conditioning of the compared methods. All decoders except LLaDA are autoregressive. $^\dagger$Run by Reg2RG with Llama-2-7B, $256 \times 256 \times 64$ volumes, and 32 visual tokens, with the volume as a multi-channel image for 2D encoders; original decoders in parentheses.}
  \label{tab:supp_baselines}
  \centering
  \footnotesize
  \setlength{\tabcolsep}{4pt}
  \begin{tabular}{@{}l p{4.3cm} p{4.4cm} p{5.7cm}@{}}
    \toprule
    Method & Visual encoder & Language decoder & Extra conditioning \\
    \midrule
    R2GenGPT$^\dagger$ & 2D Swin-B & Llama-2-7B & -- \\
    MedVInT$^\dagger$ & 2D ResNet-50 from PMC-CLIP & Llama-2-7B (PMC-LLaMA-7B) & -- \\
    RadFM$^\dagger$ & RadFM 3D ViT and Perceiver & Llama-2-7B (MedLLaMA-13B) & -- \\
    CT2Rep$^\dagger$ & 3D CT-ViT & Llama-2-7B (memory Transformer) & -- \\
    M3D$^\dagger$ & M3D-CLIP 3D ViT, spatial pooling & Llama-2-7B & -- \\
    CT-GRAPH & Frozen 3D SwinViT from VoCo & Llama-2-7B, LoRA & Graph attention over TotalSegmentator anatomy nodes \\
    DiffVP & Frozen 3D ResNet-18, Q-Former & Llama-2-7B, LoRA & Difference from a normal CT, prompt from an 18-label classifier \\
    Reg2RG & Frozen RadFM 3D ViT and Perceiver & Llama-2-7B, LoRA & Region masks, one section per region \\
    Dia-LLaMA & RadFM 3D ViT and Perceiver & Llama-2-7B, LoRA & Disease prototypes, prompt ``The \{disease\} is [state]'' \\
    \midrule
    LLaDA w/o plan & Frozen NV-Reason-CT and Perceiver & LLaDA-8B-Instruct, LoRA & Region masks, one section per region \\
    \method-AR & Frozen NV-Reason-CT and Perceiver & Meta-Llama-3-8B, LoRA & Region masks, three-state plan as text \\
    \method & Frozen NV-Reason-CT and Perceiver & LLaDA-8B-Instruct, LoRA & Region masks, three-state plan in canvas segments \\
    \bottomrule
  \end{tabular}
\end{table*}

\paragraph{Baselines.}
\Cref{tab:supp_baselines} lists the visual encoders, language decoders, and extra conditioning of all compared methods.
For the five earlier methods in \cref{tab:sota}, we copy the language metrics and CE-F1 reported by Reg2RG.
The released DiffVP predictions come from a model retrained by its authors.
Its released evaluation keeps the region prefixes ``The region $i$ is $X$:'' in both the reference and the prediction.
These shared prefixes raise BLEU-4 from 25.2 to 36.0, so we remove them.
We join the Reg2RG region sections in the fixed order of the other systems.
This raises Reg2RG's ROUGE-L from 36.7 to 38.0.
Because CT-GRAPH releases whole reports without region tags, to compute the polarity metrics we assign each generated sentence to the region of an identical training sentence, which covers 91\% of its sentences, and assign each remaining sentence to the region of the most similar training sentence under TF-IDF.
On the validation references, this assignment rule matches the region labels of RadGenome-ChestCT on 93\% of the sentences.
The sentences of Dia-LLaMA are assigned in the same way for its polarity metrics only.
On CTRG-Chest, we train Reg2RG with its official code and settings, initializing its visual encoder from RadFM.
LLaDA w/o plan is trained for 12 epochs.
\method-AR receives the plan formatted as the text prompt ``region: observation: abnormal; observation: normal; \ldots\ Report:'' and is trained to generate the sentences for the planned observations.

\paragraph{Training on CTRG-Chest.}
On CTRG-Chest, the observations, prototypes, classifiers, and plans are built from its 1,232 training CTs with the same settings.
\method starts from the first-observation model trained on RadGenome-ChestCT for 3 epochs and trains on CTRG-Chest for 6 epochs with $f = 0.3$ and E-steps before epochs 2 and 3, as on RadGenome-ChestCT.
The first-observation model in \cref{tab:supp_ext} trains in the same way without E-steps, and \method-AR starts from its RadGenome-ChestCT model and trains for 3 epochs.
The mean of GREEN and CE-F1 on the 180 validation CTs selects epoch 4 of \method and epoch 3 of the first-observation model among epochs 3 to 6, and epoch 1 of \method-AR.

\paragraph{Decoding cost.}
On 50 validation CTs with batch size 1, \method takes 12.2\,s per CT across 174 forward passes of about 70\,ms each for canvases of 154 tokens on average.
\method-AR takes 21.7\,s per CT and generates 363 tokens at about 60\,ms per token.

\section{Evaluation Details}
\label{supp:eval}

\paragraph{Significance tests.}
We compute 95\% confidence intervals from 10,000 paired bootstrap resamples over CTs.
\Cref{tab:supp_signif} lists the differences for the significance claims in the paper.

\begin{table}[t]
  \caption{Paired differences with 95\% confidence intervals from 10,000 bootstrap resamples over CTs. Each row subtracts the second system from the first, and differences are computed before rounding. Ablation rows compare entries in \cref{tab:ablation} on RadGenome-ChestCT.}
  \label{tab:supp_signif}
  \centering
  \footnotesize
  \setlength{\tabcolsep}{2.5pt}
  \begin{tabular}{@{}llcc@{}}
    \toprule
    Comparison & Metric & RadGenome & CTRG \\
    \midrule
    \multirow{2}{*}{\method $-$ Reg2RG} & GREEN & 11.9 {\scriptsize[10.7, 13.1]} & 9.6 {\scriptsize[7.8, 11.4]} \\
    & RaTE & 5.7 {\scriptsize[5.3, 6.0]} & 7.8 {\scriptsize[6.9, 8.6]} \\
    \multirow{2}{*}{\method $-$ \method-AR} & GREEN & 1.3 {\scriptsize[0.4, 2.2]} & 2.1 {\scriptsize[0.8, 3.3]} \\
    & RaTE & 1.1 {\scriptsize[0.9, 1.3]} & 2.9 {\scriptsize[2.4, 3.4]} \\
    \midrule
    \multirow{3}{*}{Learned $-$ first-obs.} & GREEN & $-$0.7 {\scriptsize[$-$1.5, 0.1]} & -- \\
    & CE-F1 & 1.0 {\scriptsize[0.5, 1.6]} & -- \\
    & FNeg & $-$3.0 {\scriptsize[$-$4.6, $-$1.5]} & -- \\
    No \Skip $-$ with \Skip & GREEN & $-$2.0 {\scriptsize[$-$2.9, $-$1.1]} & -- \\
    Write all $-$ learned start & GREEN & 0.8 {\scriptsize[$-$0.2, 1.9]} & -- \\
    \bottomrule
  \end{tabular}
\end{table}

\begin{table*}[!t]
  \caption{Results on the 1,505 RadGenome-ChestCT validation CTs released by CT-GRAPH.}
  \label{tab:supp_ctgraph}
  \centering
  \small
  \setlength{\tabcolsep}{4pt}
  \begin{tabular}{@{}l ccccc cccc ccc c@{}}
    \toprule
    Method & B-1 & B-4 & MTR & R-L & CIDEr & CE-F1 & RadGraph & GREEN & RaTE & Pos-F1 & Neg-F1 & FNeg$\downarrow$ & Words \\
    \midrule
    CT-GRAPH & 46.1 & 23.6 & 42.8 & 37.1 & \textbf{11.8} & 34.8 & 20.5 & 41.4 & 67.3 & 23.6 & 53.5 & 40.5 & 176 \\
    DiffVP & 47.9 & 24.5 & 44.5 & 37.0 & 5.3 & 38.3 & 20.0 & 33.8 & 66.9 & 24.4 & 53.6 & 34.2 & 204 \\
    Reg2RG & 46.8 & 24.2 & 43.4 & 37.3 & 8.4 & 23.9 & 20.3 & 36.8 & 66.4 & 14.6 & 53.6 & 66.7 & 181 \\
    \method & \textbf{50.8} & \textbf{29.0} & \textbf{51.8} & \textbf{40.8} & 7.1 & \textbf{60.0} & \textbf{24.8} & \textbf{48.6} & \textbf{72.0} & \textbf{43.0} & \textbf{60.9} & \textbf{21.3} & 250 \\
    \bottomrule
  \end{tabular}
\end{table*}

\paragraph{Results on the CT-GRAPH subset.}
CT-GRAPH provides predictions on 1,505 of the 1,564 validation CTs.
\Cref{tab:supp_ctgraph} evaluates all systems on these 1,505 CTs, and the ranking matches \cref{tab:sota}.

\begin{table*}[!t]
  \caption{Change in GREEN and RaTEScore per report from stating X in each way instead of leaving it unmentioned, with 95\% confidence intervals.}
  \label{tab:supp_regress}
  \centering
  \small
  \setlength{\tabcolsep}{6pt}
  \begin{tabular}{@{}lcccc@{}}
    \toprule
    & \multicolumn{2}{c}{5 systems} & \multicolumn{2}{c}{14 systems} \\
    \cmidrule(lr){2-3} \cmidrule(l){4-5}
    X, statement & GREEN & RaTE & GREEN & RaTE \\
    \midrule
    Present, ``X'' & 0.5 {\scriptsize[0.1, 0.8]} & 0.9 {\scriptsize[0.8, 1.0]} & 0.4 {\scriptsize[0.1, 0.7]} & 0.9 {\scriptsize[0.8, 0.9]} \\
    Present, ``no X'' & 1.0 {\scriptsize[0.2, 2.0]} & 0.6 {\scriptsize[0.3, 0.8]} & 0.6 {\scriptsize[$-$0.1, 1.3]} & 0.5 {\scriptsize[0.4, 0.7]} \\
    Absent, ``no X'' & 2.5 {\scriptsize[2.3, 2.8]} & 1.0 {\scriptsize[1.0, 1.1]} & 2.4 {\scriptsize[2.3, 2.5]} & 1.0 {\scriptsize[0.9, 1.0]} \\
    Absent, ``X'' & $-$1.3 {\scriptsize[$-$1.5, $-$1.1]} & $-$0.3 {\scriptsize[$-$0.3, $-$0.2]} & $-$1.1 {\scriptsize[$-$1.3, $-$1.0]} & $-$0.2 {\scriptsize[$-$0.3, $-$0.2]} \\
    \bottomrule
  \end{tabular}
\end{table*}

\paragraph{Metric regression.}
\Cref{tab:supp_regress} reports the regression coefficients plotted in \cref{fig:analysis}b alongside the same regression fit across all 14 validation systems totaling 21,896 reports.
Each coefficient estimates the change in a score per report when observation X is stated in a given way rather than left unmentioned.
A false negation raises GREEN by 1.0 $[0.2, 2.0]$, no less than stating a present X, which raises it by 0.5 $[0.1, 0.8]$.
The regression uses one intercept per CT and one per system, with confidence intervals from 1,000 bootstrap resamples over CTs.
The five systems in the main regression are \method, the first-observation model, \method-AR, LLaDA w/o plan, and Reg2RG.
The 14 systems add the first-observation model at $\kappa = 0.5$ and $0.4$ and with positives only, \method at $\kappa = 0.4$, \method-AR at $\kappa = 0.5$, the model without \Skip targets in \cref{tab:ablation}, and the three other start rules in \cref{tab:supp_start}.
RaTEScore is not computed for the positives-only plan, so its 14-system fit uses the other 13 systems.
GREEN's own error counts agree, because a false negation changes false findings by $-0.16$ $[-0.30, -0.04]$ and missed findings by $-0.08$ $[-0.19, 0.03]$ relative to leaving X unmentioned, whereas a false positive adds 0.21 false findings.

\paragraph{Controlled edits.}
We apply controlled edits to the reference reports of 504 validation CTs to test each metric without generated text.
In each CT, we select one sentence that states a single present observation not stated by any other sentence in the CT.
The observation is one of the 18 CE labels whose most frequent training sentence stating only ``no X'' occurs at least 20 times.
This criterion leaves lung nodule, lymphadenopathy, pericardial effusion, pleural effusion, and consolidation.
We either remove the sentence so that X is left unmentioned or replace it with that ``no X'' sentence to introduce a false negation.
\Cref{tab:supp_edits} reports the scores of both edits across all 1,564 CTs.
CE-F1, RadGraph F1, and RaTEScore score the two edits nearly the same.
Per edited CT, the RaTEScore of the false negation differs from that of leaving X unmentioned by $-0.03$ $[-0.13, 0.07]$.
GREEN instead penalizes the false negation by 3.5 $[2.7, 4.3]$ more per edited CT.
GREEN therefore detects the contradiction in these clean edits, unlike on the generated reports of \cref{fig:analysis}b.
The polarity metrics also separate the edits.
Negative F1 falls to 98.9, and false negations rise to 32.2 per 100 reports.

\begin{table}[t]
  \caption{Reference reports of the 1,564 validation CTs, with one present finding left unmentioned or stated as absent in 504 reports.}
  \label{tab:supp_edits}
  \centering
  \footnotesize
  \setlength{\tabcolsep}{1.4pt}
  \begin{tabular}{@{}lccccccc@{}}
    \toprule
    Report & CE-F1 & RadGraph & GREEN & RaTE & Pos-F1 & Neg-F1 & FNeg$\downarrow$ \\
    \midrule
    Reference & 99.9 & 99.6 & 97.7 & 99.9 & 100.0 & 100.0 & 0.0 \\
    X unmentioned & 93.7 & 92.5 & 96.6 & 98.9 & 97.2 & 100.0 & 0.0 \\
    X denied & 93.7 & 92.7 & 95.5 & 98.9 & 97.2 & 98.9 & 32.2 \\
    \bottomrule
  \end{tabular}
\end{table}

\paragraph{Report-level polarity metrics.}
The polarity metrics in the main paper match observations within each region section.
For CT-GRAPH and Dia-LLaMA, whose sentences are assigned to regions as described in \cref{supp:impl}, a sentence placed in a wrong region therefore counts as both a missed and an unsupported statement.
\Cref{tab:supp_polarity_report} removes this dependence by pooling the observations of all region sections of a CT before matching, taking the positive state when an observation is stated in both states.
With pooling, every prior method still trails \method on all three polarity metrics, and \method-AR, which receives the same plans, stays within 1.4 of \method.

\begin{table}[t]
  \caption{Report-level polarity metrics, with observations pooled over the region sections of each CT, on the 1,505 RadGenome-ChestCT validation CTs released by CT-GRAPH and the CTRG-Chest test set.}
  \label{tab:supp_polarity_report}
  \centering
  \footnotesize
  \setlength{\tabcolsep}{3pt}
  \begin{tabular}{@{}llccc@{}}
    \toprule
    Data & Method & Pos-F1 & Neg-F1 & FNeg$\downarrow$ \\
    \midrule
    \multirow{5}{*}{RadGenome} & CT-GRAPH & 25.1 & 56.5 & 45.2 \\
    & DiffVP & 26.6 & 56.5 & 36.7 \\
    & Reg2RG & 14.4 & 57.1 & 74.2 \\
    & \method-AR & 45.3 & \textbf{63.6} & \textbf{21.9} \\
    & \method & \textbf{46.0} & 63.5 & 23.3 \\
    \midrule
    \multirow{4}{*}{CTRG} & Dia-LLaMA & 29.2 & 63.2 & 29.3 \\
    & Reg2RG & 23.6 & 56.7 & 40.3 \\
    & \method-AR & 40.3 & 65.5 & \textbf{19.9} \\
    & \method & \textbf{40.5} & \textbf{66.0} & 21.0 \\
    \bottomrule
  \end{tabular}
\end{table}

\paragraph{CIDEr and report length.}
CIDEr-D scales each $n$-gram cosine similarity by $e^{-\delta^2/(2\sigma^2)}$, where $\delta$ is the length difference between the generated and reference reports and $\sigma = 6$.
For reports of about 200 words, this Gaussian length factor dominates the metric, as 67\% to 74\% of the CIDEr score of each system in \cref{tab:supp_cider} comes from the 5.7\% to 8.1\% of CTs whose generated report length falls within 6 words of the reference.
\Cref{tab:supp_cider} recomputes CIDEr without the length penalty, and \method achieves the highest unpenalized CIDEr on both sets of CTs.
Writing every segment lowers unpenalized CIDEr from 96.1 to 79.5 for the first-observation model, which shows that repeated sentences also reduce $n$-gram matches.

\begin{table}[t]
  \caption{CIDEr with and without the length penalty. ``Within 6'' is the share of CTs whose report length is within 6 words of the reference, which averages 194 and 196 words on the two sets.}
  \label{tab:supp_cider}
  \centering
  \small
  \setlength{\tabcolsep}{3pt}
  \begin{tabular}{@{}llcccc@{}}
    \toprule
    CTs & Method & CIDEr & No penalty & Words & Within 6 \\
    \midrule
    \multirow{3}{*}{1,564} & Reg2RG & 9.1 & 66.6 & 181 & 8.1\% \\
    & DiffVP & 5.8 & 61.9 & 203 & 7.9\% \\
    & \method & 6.8 & \textbf{100.1} & 247 & 5.7\% \\
    \midrule
    \multirow{4}{*}{1,505} & CT-GRAPH & 11.8 & 70.8 & 176 & 7.5\% \\
    & Reg2RG & 8.4 & 62.3 & 181 & 8.0\% \\
    & DiffVP & 5.3 & 58.2 & 204 & 7.8\% \\
    & \method & 7.1 & \textbf{93.9} & 250 & 5.9\% \\
    \bottomrule
  \end{tabular}
\end{table}

\section{Additional Results}
\label{supp:results}

\begin{figure*}[!t]
  \centering
  \includegraphics[width=\linewidth]{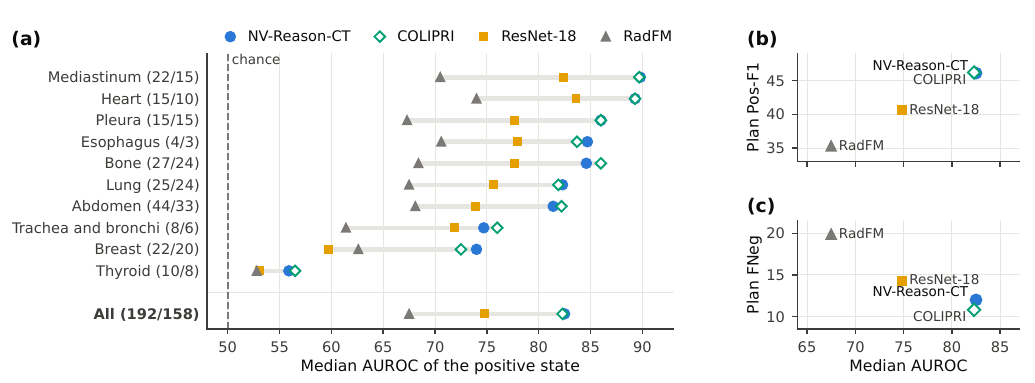}
  \caption{Classifiers and plans of the four frozen encoders. (a)~Median out-of-fold AUROC of the positive state per region, with the numbers of observations and finding classifiers after each region. (b)~Positive F1 and (c)~false negations per 100 reports of the plans on all 1,564 validation CTs against the median AUROC over the 158 finding classifiers.}
  \label{fig:supp_classifiers}
\end{figure*}

\paragraph{Classifier quality.}
\Cref{fig:supp_classifiers}a reports, for each region, the number of observations, the number with a finding classifier, and the median AUROC of the positive state, computed out of fold with 5 folds grouped by patient over the training CTs.
Besides NV-Reason-CT, it covers the three frozen encoders described in the next paragraph.
NV-Reason-CT and COLIPRI achieve nearly the same AUROC in every region, with medians of 82.5 and 82.3 over all 158 finding classifiers, whereas RadFM is 15.0 lower and the 3D ResNet-18 is 7.7 lower.
We attribute the near-chance thyroid classifiers on every encoder to the small size of the thyroid, which covers on average only 1.1 grid cells by at least half in \cref{tab:supp_masks}.
On the 230 development CTs, the region gate passes 8.09 of the 9.86 masked regions per CT, with a precision of 92.0 and a recall of 99.4 for the regions written in the reference.

\begin{table*}[!t]
  \caption{Generators with different frozen encoders on RadGenome-ChestCT validation. AR w/o plan uses Meta-Llama-3-8B with NV-Reason-CT and RadLLaMA-7B with RadFM.}
  \label{tab:supp_encoders}
  \centering
  \small
  \setlength{\tabcolsep}{6pt}
  \begin{tabular}{@{}llccccccc@{}}
    \toprule
    Encoder & Generator & Words & CE-F1 & GREEN & RaTE & Pos-F1 & Neg-F1 & FNeg$\downarrow$ \\
    \midrule
    \multirow{4}{*}{NV-Reason-CT} & AR w/o plan & 169 & 45.7 & 50.0 & 71.1 & 36.1 & 59.7 & 42.5 \\
    & LLaDA w/o plan & 231 & 37.7 & 37.6 & 69.6 & 28.8 & 60.0 & 30.0 \\
    & \method-AR & 240 & 57.2 & 48.3 & 71.4 & 42.3 & 61.6 & 19.4 \\
    & \method & 247 & 59.7 & 49.6 & 72.5 & 42.9 & 61.6 & 20.5 \\
    \midrule
    \multirow{2}{*}{COLIPRI} & LLaDA w/o plan & 241 & 32.4 & 34.9 & 69.1 & 26.9 & 59.9 & 32.2 \\
    & \method & 221 & 57.6 & 48.6 & 72.8 & 41.8 & 61.3 & 23.8 \\
    \midrule
    \multirow{3}{*}{RadFM} & AR w/o plan & 152 & 32.5 & 45.8 & 68.6 & 23.4 & 57.9 & 45.8 \\
    & LLaDA w/o plan & 219 & 29.4 & 39.9 & 70.6 & 30.0 & 60.0 & 50.4 \\
    & \method & 190 & 41.9 & 44.2 & 72.1 & 34.3 & 60.0 & 37.3 \\
    \bottomrule
  \end{tabular}
\end{table*}

\paragraph{Other visual encoders.}
\Cref{tab:supp_encoders} replaces NV-Reason-CT with two other frozen 3D encoders and trains \method with the same recipe, using the classifiers, region gate, plans, and segment lengths of each encoder.
COLIPRI is a Primus encoder pretrained on CT-RATE and NLST with contrastive, report generation, and masked image objectives, and its tokens lie on the same $24^3$ grid as those of NV-Reason-CT.
RadFM~\cite{radfm} is the frozen encoder of Reg2RG, so its visual prefix follows Reg2RG, and its region feature joins the mean token inside the region mask with the mean token of the volume.
We also train classifiers and build plans on a 3D ResNet-18 trained on CT-RATE with the contrastive objective of CT-CLIP~\cite{ctrate}, but we do not train a generator on it.

\paragraph{Plans with other encoders.}
\Cref{fig:supp_classifiers}b and c show that plan quality follows classifier quality.
NV-Reason-CT and COLIPRI yield plans with a positive F1 of 46.1 and 46.2 and with 12.0 and 10.8 false negations per 100 reports, whereas RadFM yields a positive F1 of 35.3 and 19.9 false negations.
The ResNet-18 lies between them on both measures.

\paragraph{Reports with other encoders.}
After generation, \method with COLIPRI stays close to \method with NV-Reason-CT, with a GREEN of 48.6 and a CE-F1 of 57.6 compared with 49.6 and 59.7.
Its reports, however, contain 23.8 false negations per 100 reports compared with 20.5, although its plans contain fewer.
With RadFM, CE-F1 falls to 41.9 and false negations rise to 37.3, in line with its weaker classifiers.
Even with the frozen RadFM encoder of Reg2RG, \method raises CE-F1 from 23.7 to 41.9 and positive F1 from 14.4 to 34.3 over Reg2RG in \cref{tab:sota} and makes 42\% fewer false negations.
The plan improves the generator with every encoder.
Without a plan, LLaDA reaches a CE-F1 of 37.7 with NV-Reason-CT and 29.4 with RadFM, and the plan raises it to 59.7 and 41.9 while reducing false negations by 32\% and 26\%.
With COLIPRI, the plan raises CE-F1 from 32.4 to 57.6 and positive F1 from 26.9 to 41.8, and likewise reduces false negations by 26\%.
These results suggest that the encoder affects \method mainly through its classifiers.

\paragraph{AR generators without a plan.}
\Cref{tab:supp_encoders} also includes AR generators without a plan.
With NV-Reason-CT, AR w/o plan is trained as \method-AR but without the plan in its prompt, and with RadFM, it follows LLaDA w/o plan with RadLLaMA-7B in place of LLaDA-8B-Instruct.
Their GREEN is 50.0 with NV-Reason-CT and 45.8 with RadFM, compared with 49.6 and 44.2 for \method, and only the gap of 1.6 $[0.5, 2.7]$ with RadFM is significant.
They state only 3.8 and 2.9 positives per report, compared with 6.8 and 5.0 for \method, so they make fewer false positive statements, which GREEN penalizes.
They also miss more findings, with a lower CE-F1, positive F1, and RaTEScore, and make 2.1 and 1.2 times as many false negations as \method, which GREEN does not penalize, as \cref{sec:analysis} shows.
With NV-Reason-CT, the plan raises the CE-F1 of the AR generator from 45.7 to 57.2 and lowers its false negations from 42.5 to 19.4, while its GREEN falls from 50.0 to 48.3.

\paragraph{Rules at equal negatives.}
\Cref{tab:supp_neg_rules} compares rules that choose the planned negatives from the same three-state probabilities, region gate, and planned positives on all 1,564 validation CTs.
The single threshold plans an observation as negative when $p_{i1} > 0.2545$, and the top-N rule plans the N observations with the highest $p_{i1}$ in each region section, where N is the number of negatives that the expected-cost rule plans in that section.
Both rules thus plan as many negatives as the expected-cost rule.
The three-state argmax plans the most probable state of each observation and states only 13.7 negatives and 3.0 positives per report.
At an equal number of negatives, the expected-cost rule plans 12.0 false negations per 100 reports, compared with 20.3 for the single threshold and 17.5 for the top-N rule, whereas the negative F1 of the three rules differs by at most 0.4.
The expected-cost rule withholds a negative when $p_{i2}$ is high relative to $p_{i1}$, which the other two rules do not consider.
We decode each plan with the final generator and predicted segment lengths.
The expected-cost rule then makes 5.4 $[4.2, 6.6]$ fewer false negations per 100 reports than the single threshold and 1.7 $[0.8, 2.7]$ fewer than the top-N rule.
GREEN, CE-F1, and RaTEScore differ by at most 0.3 among the three rules, and GREEN is 49.6 for both the expected-cost rule and the single threshold, so these metrics do not reveal the difference, as \cref{sec:analysis} discusses.
Generation adds false negations to every plan, from 12.0 planned to 20.5 written for the expected-cost rule, and we trace their sources below.

\begin{table*}[!t]
  \caption{Rules for planned negatives on all 1,564 validation CTs, scored on the plans and on the reports of the final generator. Neg.\ is the number of planned negatives per report.}
  \label{tab:supp_neg_rules}
  \centering
  \small
  \setlength{\tabcolsep}{6pt}
  \begin{tabular}{@{}lc cc ccccc@{}}
    \toprule
    & & \multicolumn{2}{c}{Plan} & \multicolumn{5}{c}{Report} \\
    \cmidrule(lr){3-4} \cmidrule(l){5-9}
    Rule & Neg. & Neg-F1 & FNeg$\downarrow$ & GREEN & CE-F1 & RaTE & Neg-F1 & FNeg$\downarrow$ \\
    \midrule
    Expected-cost rule & 23.4 & 61.9 & 12.0 & 49.6 & 59.7 & 72.5 & 61.6 & 20.5 \\
    Single threshold on $p_{i1}$ & 23.3 & 62.3 & 20.3 & 49.6 & 59.5 & 72.4 & 61.9 & 25.9 \\
    Top-N per section & 23.4 & 62.1 & 17.5 & 49.3 & 59.6 & 72.4 & 61.8 & 22.3 \\
    Three-state argmax & 13.7 & 60.0 & 21.9 & -- & -- & -- & -- & -- \\
    \bottomrule
  \end{tabular}
\end{table*}

\begin{figure}[t]
  \centering
  \includegraphics[width=\linewidth]{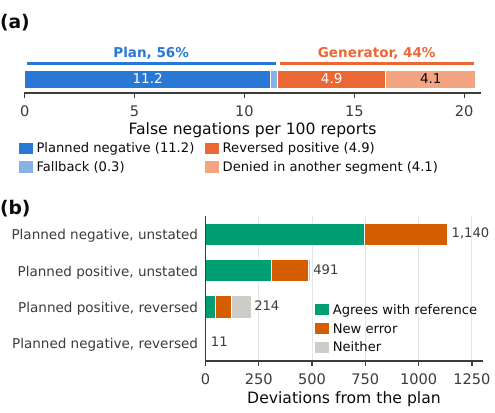}
  \caption{Sources of false negations and deviations from the plan of \method on the 1,564 validation CTs. (a)~False negations per 100 reports caused by the plan or by the generator. (b)~Deviations of each type, split by whether the section state agrees with the reference, introduces an error, or differs from both.}
  \label{fig:supp_fneg}
\end{figure}

\paragraph{Deviations from the plan.}
The generator states the planned state in 91.8\% of the written positive segments and 99.9\% of the written negative segments.
We count a deviation when the final state of an observation in its region section differs from its planned state, including a planned observation that the section leaves unstated.
Of the 1,856 deviations on the validation CTs, 59.5\% agree with the reference, 34.4\% introduce an error, and the rest differ from both, as \cref{fig:supp_fneg}b shows by type.
Most deviations leave a planned observation unstated, which happens to 1,140 planned negatives and 491 planned positives, and the reference does not state 65\% and 64\% of them either.
The generator reverses 214 planned positives to negative, and the reference states 77 of them as positive.

\paragraph{Sources of false negations.}
Of the 20.5 false negations per 100 reports, the plan accounts for 56\%, as \cref{fig:supp_fneg}a shows.
Planned negatives that the reference states as positive contribute 11.2, mostly for mediastinal lymphadenopathy and pericardial effusion, and the fallback for regions without a planned positive contributes 0.3.
The generator accounts for the other 44\%, with 4.9 from reversed planned positives and 4.1 from sentences in the segment of another observation that deny an unplanned one, mostly lung nodules.

\begin{figure*}[!t]
  \centering
  \includegraphics[width=\linewidth]{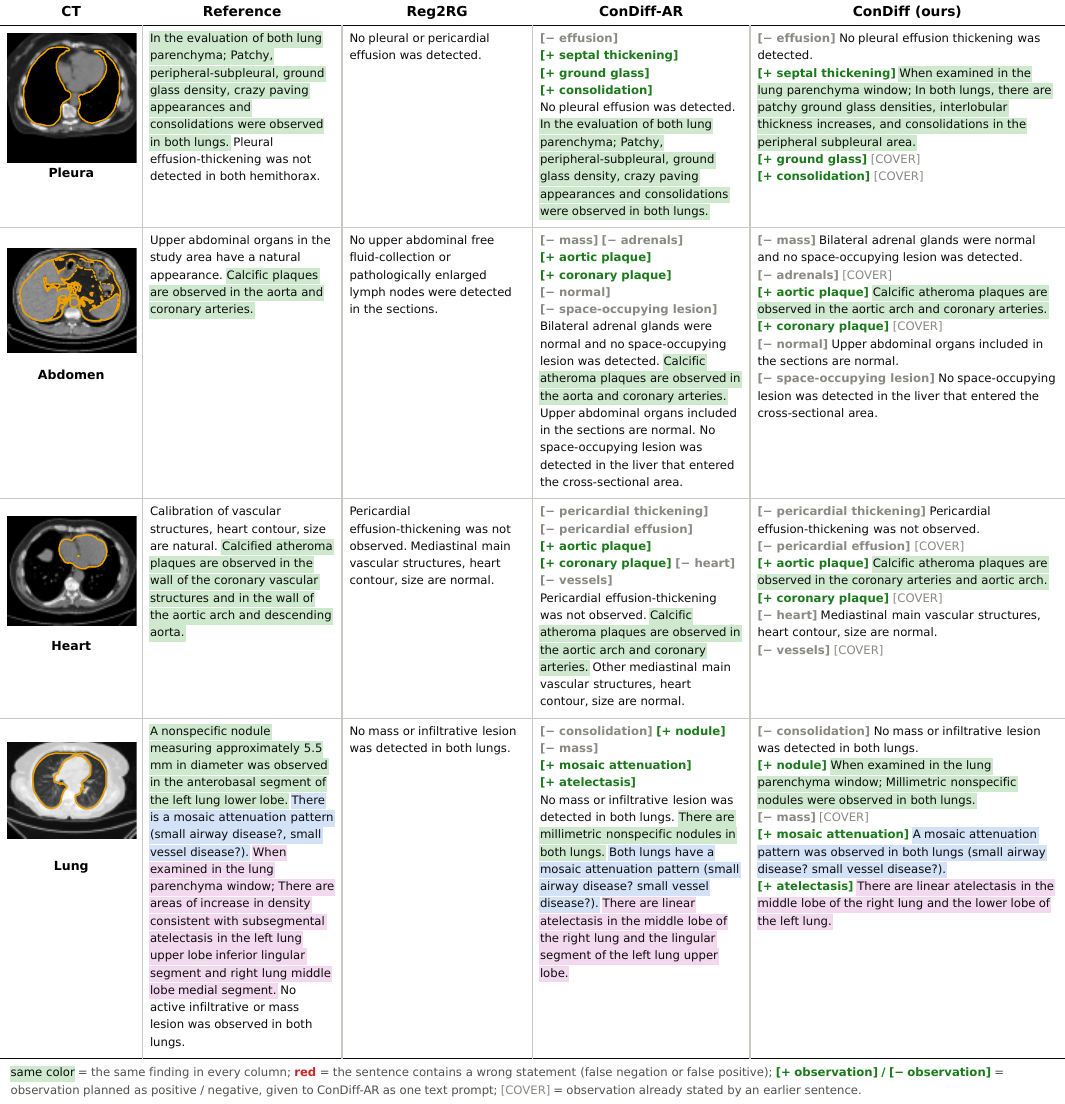}
  \caption{More region sections of RadGenome-ChestCT validation CTs, drawn as in \cref{fig:cases}.}
  \label{fig:supp_cases}
\end{figure*}

\paragraph{Dependence on the visual prefix.}
\Cref{tab:supp_visual} decodes 200 randomly drawn validation CTs with their own plans and segment lengths while changing only the visual prefix.
With the prefix of another patient's CT in the same region, 88.3\% of the region sections stay unchanged, and with a zero prefix, 66.2\% do.
CE-F1, GREEN, RaTEScore, positive F1, and negative F1 change by at most 0.4, and false negations change by at most 1.5 per 100 reports.
The plan thus determines most of each report.
The image reaches the report mainly through the classifiers that build the plan from the same frozen encoder: without a plan, the same generator reads the image only through its visual prefix and reaches a CE-F1 of 37.7, compared with 59.7 for \method (\cref{tab:supp_encoders}).
Because the plan holds no sizes or segment-level locations, \method states them only in general terms, as in the lung case of \cref{fig:supp_cases}.

\begin{table*}[!t]
  \caption{The final generator on 200 validation CTs with the visual prefix of the same CT, of another patient's CT, or of zero vectors. Same is the share of unchanged region sections.}
  \label{tab:supp_visual}
  \centering
  \small
  \setlength{\tabcolsep}{6pt}
  \begin{tabular}{@{}lcccccccc@{}}
    \toprule
    Visual prefix & Same (\%) & Words & CE-F1 & GREEN & RaTE & Pos-F1 & Neg-F1 & FNeg$\downarrow$ \\
    \midrule
    Same CT & 100 & 241 & 60.7 & 49.8 & 72.1 & 42.7 & 61.1 & 27.0 \\
    Another patient & 88.3 & 241 & 60.5 & 49.8 & 72.2 & 42.8 & 61.1 & 28.0 \\
    Zero & 66.2 & 242 & 61.1 & 49.9 & 72.2 & 42.5 & 61.1 & 25.5 \\
    \bottomrule
  \end{tabular}
\end{table*}

\begin{table*}[!t]
  \caption{Plan rules with the final generator and $\lambda = 4$ on all 1,564 validation CTs. Neg.\ is the number of negatives stated per generated report, whereas \cref{tab:supp_neg_rules} counts planned negatives. The $\kappa = 0.3$ row is \method, and the two-state row is (2) in \cref{tab:ablation}.}
  \label{tab:supp_kappa}
  \centering
  \small
  \setlength{\tabcolsep}{6pt}
  \begin{tabular}{@{}lccccccc@{}}
    \toprule
    Plan & Neg. & Words & B-4 & CE-F1 & GREEN & Neg-F1 & FNeg$\downarrow$ \\
    \midrule
    Reference & 14.8 & -- & -- & -- & -- & -- & -- \\
    Positives only & 7.5 & 146 & 19.3 & 59.8 & 37.1 & 45.9 & 10.1 \\
    $\kappa = 0.5$ & 19.6 & 226 & 29.1 & 60.2 & 45.9 & 62.5 & 18.4 \\
    $\kappa = 0.4$ & 21.2 & 237 & 29.5 & 59.9 & 47.8 & 62.4 & 19.6 \\
    $\kappa = 0.3$ & 23.2 & 247 & 29.6 & 59.7 & 49.6 & 61.6 & 20.5 \\
    $\kappa = 0$ & 51.2 & 498 & 15.9 & 39.9 & 42.4 & 37.8 & 41.7 \\
    Two-state & 52.8 & 1108 & 8.8 & 7.1 & 17.3 & 34.2 & 52.8 \\
    \bottomrule
  \end{tabular}
\end{table*}

\paragraph{Plan rule.}
\Cref{tab:supp_kappa} varies the plan rule for the final generator on all 1,564 validation CTs, as in \cref{fig:analysis}a.
Lowering $\kappa$ from 0.5 to 0.3 adds negatives and raises GREEN while negative F1 stays between 61.6 and 62.5, whereas planning only positives, $\kappa = 0$, and the two-state rule all lower GREEN.

\paragraph{CE-F1 truncation.}
Because the CE-F1 classifier truncates its input at 512 tokens, it reads on average only 34\% of a two-state report, which averages 1,526 tokens, and 69\% of a report at $\kappa = 0$.
In contrast, it reads nearly all reports of \method at $\kappa = 0.3$ and all reference reports in full.
The CE-F1 of the two-state rule and of $\kappa = 0$ in \cref{tab:supp_kappa} thus mostly reflects truncation, whereas false negations and negative F1 count every statement without truncation.

\paragraph{Choice of $\kappa$.}
We check our choice of $\kappa$ on the 230 development CTs with the final generator and predicted segment lengths.
We score each value by the mean of GREEN and CE-F1, a rule fixed before decoding.
This mean is 54.1 at $\kappa = 0.5$, 55.0 at $\kappa = 0.4$, 54.8 at $\kappa = 0.3$, and 42.7 at $\kappa = 0$.
The development CTs cannot separate $\kappa = 0.3$ from $\kappa = 0.4$, so we keep $\kappa = 0.3$.
On all 1,564 validation CTs, at $\kappa = 0.5$ and $\kappa = 0.4$, the final generator outperforms all prior methods in \cref{tab:sota} on every metric except CIDEr.
GREEN is 45.9 and 47.8, with 18.4 and 19.6 false negations per 100 reports, respectively.

\begin{figure}[H]
  \centering
  \includegraphics[width=\linewidth]{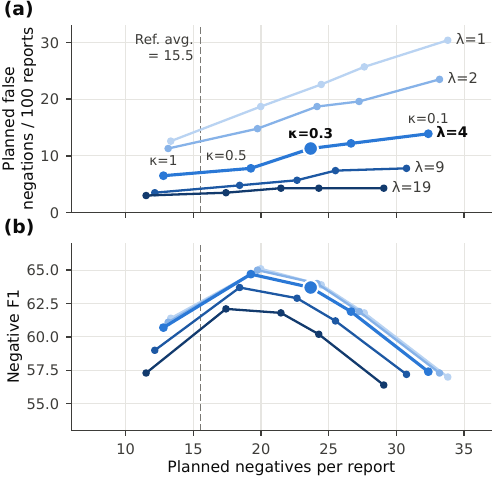}
  \caption{Plans of the 230 development CTs before generation as $\lambda$ and $\kappa$ vary. Each curve fixes $\lambda$ and passes through $\kappa = 1$, 0.5, 0.3, 0.2, and 0.1 from left to right. (a) Planned false negations per 100 reports. (b) Negative F1 of the plan. The large dot marks $\lambda = 4$ and $\kappa = 0.3$, and the dashed line marks the mean number of negatives per reference report.}
  \label{fig:supp_lambda_kappa}
\end{figure}

\paragraph{Plans as $\lambda$ and $\kappa$ vary.}
\Cref{fig:supp_lambda_kappa} shows how $\lambda$ and $\kappa$ affect the plans of the 230 development CTs before generation.
A larger $\kappa$ yields fewer planned negatives, and a larger $\lambda$ yields fewer planned false negations.
At $\kappa = 0.3$, raising $\lambda$ from 1 to 4 lowers planned false negations from 22.6 to 11.3 per 100 reports.
Across $\kappa$ from 0.1 to 1, negative F1 differs by at most 0.7 among $\lambda = 1$, 2, and 4, whereas $\lambda = 9$ and $\lambda = 19$ have lower negative F1 than $\lambda = 4$ at every such $\kappa$.
Selecting $\lambda$ by GREEN would instead favor false negations, which GREEN rewards on generated reports (\cref{sec:analysis}).

\paragraph{Segment start rule.}
\Cref{tab:supp_start} compares the learned start rule $q^{\text{write}}_k > q^{\text{cov}}_k$ of the first-observation model with three rules for the first-observation model trained without \Cov.
The first rule writes every segment.
The merge rule leaves the segment of a planned negative $j$ empty when an earlier written segment $h$ usually states $j$ within its own sentence.
It uses pairs that at least 30 training region sections state together, with $h$ in its planned state and $j$ negative, and requires that in at least half of these sections the first sentence stating $j$ belongs to $h$ under the first-observation assignment.
The share threshold leaves segment $k$ empty when $q^{\text{write}}_k / (q^{\text{write}}_k + q^{\eos}_k) < 0.3$ at its first position.
Under every rule, the last segment of a region is written if no earlier segment is written.
\Cref{tab:supp_signif} shows that writing every segment does not significantly change GREEN, yet it lowers BLEU-4 by 4.8 and CIDEr from 9.4 to 1.5.
The merge rule lowers GREEN by 0.6, and the share threshold lowers it by 6.7.

\begin{table}[H]
  \caption{Segment start rules of the first-observation model on all validation CTs.}
  \label{tab:supp_start}
  \centering
  \footnotesize
  \setlength{\tabcolsep}{2.5pt}
  \begin{tabular}{@{}lcccccc@{}}
    \toprule
    Start rule & Words & B-4 & CIDEr & CE-F1 & GREEN & RaTE \\
    \midrule
    $q^{\text{write}} > q^{\text{cov}}$ & 221 & 29.9 & 9.4 & 58.7 & 50.2 & 72.9 \\
    Write every segment & 293 & 25.1 & 1.5 & 59.0 & 51.1 & 73.6 \\
    Merge rule & 243 & 29.2 & 7.2 & 59.3 & 49.6 & 72.9 \\
    Share threshold 0.3 & 207 & 28.3 & 9.4 & 59.2 & 43.5 & 71.7 \\
    \bottomrule
  \end{tabular}
\end{table}

\begin{table*}[!t]
  \caption{Positive recall (R, \%) and false negations per 100 reports (FNeg) of the five observations with the most false negations, on the same 1,505 CTs. $n$ is the number of positive statements in the reference region sections.}
  \label{tab:supp_fneg_obs}
  \centering
  \small
  \setlength{\tabcolsep}{4pt}
  \begin{tabular}{@{}lc cc cc cc cc@{}}
    \toprule
    & & \multicolumn{2}{c}{CT-GRAPH} & \multicolumn{2}{c}{DiffVP} & \multicolumn{2}{c}{Reg2RG} & \multicolumn{2}{c}{\method} \\
    \cmidrule(lr){3-4} \cmidrule(lr){5-6} \cmidrule(lr){7-8} \cmidrule(l){9-10}
    Observation & $n$ & R & FNeg$\downarrow$ & R & FNeg$\downarrow$ & R & FNeg$\downarrow$ & R & FNeg$\downarrow$ \\
    \midrule
    Mediastinal lymphadenopathy & 334 & 12.0 & 14.6 & 15.0 & 12.2 & 2.4 & 17.1 & \textbf{48.2} & \textbf{5.6} \\
    Lung nodule & 699 & 37.8 & 7.0 & \textbf{67.2} & \textbf{0.5} & 16.7 & 19.5 & 66.1 & 3.1 \\
    Consolidation & 301 & 22.3 & 4.9 & 17.6 & 3.4 & 7.0 & 9.9 & \textbf{66.1} & \textbf{1.1} \\
    Pleural effusion & 168 & 33.3 & 3.7 & 35.1 & 5.4 & 12.5 & 7.7 & \textbf{78.0} & \textbf{1.3} \\
    Pericardial effusion & 117 & 5.1 & \textbf{4.3} & 14.5 & 4.9 & 0.9 & 4.8 & \textbf{26.5} & 5.3 \\
    \bottomrule
  \end{tabular}
\end{table*}

\paragraph{Learned assignment.}
Compared with the first-observation assignment, the learned assignment adds 85 true positives across the 18 CE labels, 64 of them for lymphadenopathy.
Among the 439 segments planned as positive for mediastinal lymph nodes, the first-observation model writes only a negative such as ``No pathological LAP was detected in the mediastinum'' in 170 segments, or 39\%, compared with 54 segments, or 12\%, under the learned assignment.
Lymphadenopathy is recovered only by the learned assignment in 71 CTs and only by the first-observation assignment in 7 CTs.
This gain does not come from writing more positive segments, because the fraction of positive segments that start as \Cov is 13.8\% for the first-observation model and 15.1\% for the learned assignment.
In GREEN's counts, the learned assignment matches 0.18 fewer findings per report while making 0.07 fewer errors, so GREEN does not change significantly.

\begin{figure*}[!t]
  \centering
  \includegraphics[width=\linewidth]{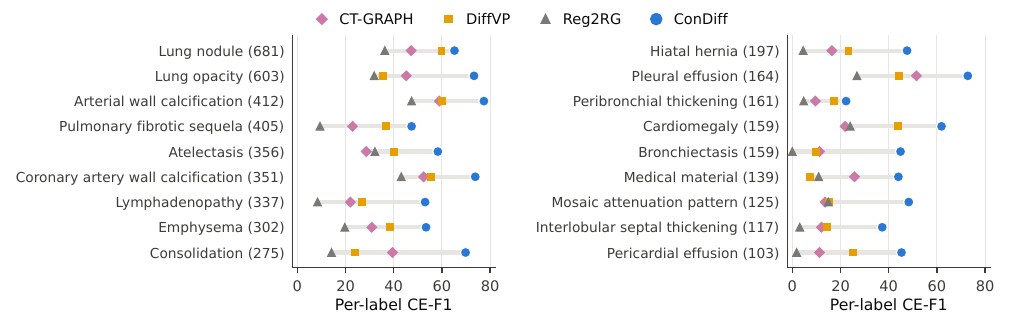}
  \caption{Per-label CE-F1 on the 1,505 RadGenome-ChestCT validation CTs released by CT-GRAPH. The number after each label counts the CTs whose reference report the CT-RATE classifier labels positive.}
  \label{fig:supp_labels}
\end{figure*}

\paragraph{Results per finding.}
\Cref{fig:supp_labels} breaks down CE-F1 across the 18 CT-RATE labels on the 1,505 CTs of \cref{tab:supp_ctgraph}.
CE-F1 is the mean of the per-label F1 scores weighted by the number of CTs whose reference report the classifier labels positive.
\method achieves the highest F1 on all 18 labels.
Its gains over the best prior method range from 5.1 on peribronchial thickening and 5.4 on lung nodules to 33.6 on bronchiectasis.
We report only RadGenome-ChestCT because the CTRG-Chest test set has fewer than 50 positive CTs for 16 of the 18 labels.

\Cref{tab:supp_fneg_obs} reports positive recall and false negations for the five observations with the most false negations.
These five account for 77\% to 89\% of the false negations of each system.
On mediastinal lymphadenopathy, \method raises recall from at most 15.0\% to 48.2\% and reduces false negations from 12.2--17.1 to 5.6 per 100 reports.
Consolidation and pleural effusion follow the same pattern.
On lung nodules, DiffVP reaches 67.2\% recall and 0.5 false negations, whereas \method reaches 66.1\% and 3.1.
Pericardial effusion is the exception.
\method still raises recall from at most 14.5\% to 26.5\% but makes 5.3 false negations per 100 reports, slightly above the 4.3 to 4.9 of prior methods.

\paragraph{Segments and generators.}
\Cref{tab:supp_canvas} separates the observation canvas from the generator on RadGenome-ChestCT validation.
LLaDA with a text plan receives the plan of \method as the text prompt of \method-AR and has no position constraints.
It writes the planned sentences of a region in canonical order on one canvas whose length is the sum of the predicted segment lengths.
It is trained for three epochs on the canvases of the first-observation model joined into one segment, using the same visual prefix and LoRA.
Its CE-F1 is 47.8, compared with 58.7 for the first-observation model.
Its GREEN and RaTEScore are lower by 3.7 $[2.7, 4.7]$ and 2.4 $[2.2, 2.6]$.
After two epochs, its CE-F1 is 50.1, so the gap does not come from the training length.
Segmented AR trains \method-AR with a batch size of 16 on the canvas of \method written as text.
This text canvas contains one line per planned observation, holding the sentences assigned to it by the learned assignment, the word ``covered'', or nothing.
Segments raise the GREEN of the AR generator by 1.2 $[0.3, 2.1]$ and its RaTEScore by 0.7 $[0.5, 0.8]$.
With segments, \method and segmented AR reach a similar GREEN, differing by 0.1 $[-0.7, 1.0]$.
However, \method keeps a higher RaTEScore by 0.5 $[0.3, 0.6]$, a higher CE-F1, and fewer false negations at 20.5 compared with 23.3 per 100 reports.
Segmented AR also relies on the assignment that \method learns by hard EM.
Most of the GREEN gain of \method over \method-AR thus comes from the observation canvas.

\begin{table}[H]
  \caption{Observation canvas and generator on RadGenome-ChestCT validation. All rows receive the same plans.}
  \label{tab:supp_canvas}
  \centering
  \footnotesize
  \setlength{\tabcolsep}{1.7pt}
  \begin{tabular}{@{}lccccccc@{}}
    \toprule
    Model & Words & CE-F1 & GREEN & RaTE & Pos-F1 & Neg-F1 & FNeg$\downarrow$ \\
    \midrule
    \method & 247 & 59.7 & 49.6 & 72.5 & 42.9 & 61.6 & 20.5 \\
    First-observation & 221 & 58.7 & 50.2 & 72.9 & 43.2 & 61.0 & 23.5 \\
    LLaDA, text plan & 173 & 47.8 & 46.6 & 70.5 & 35.1 & 57.9 & 28.1 \\
    \midrule
    \method-AR & 240 & 57.2 & 48.3 & 71.4 & 42.3 & 61.6 & 19.4 \\
    Segmented AR & 222 & 58.7 & 49.4 & 72.1 & 43.2 & 61.3 & 23.3 \\
    \bottomrule
  \end{tabular}
\end{table}

\paragraph{Generators on CTRG-Chest.}
\label{supp:external}
\Cref{tab:supp_ext} extends \cref{tab:mdlm_ar} on CTRG-Chest with the first-observation model.
The first-observation model obtains a higher positive F1 than \method, 36.4 compared with 33.0.
The E-step also reassigns 12.3\% of the training sentences on CTRG-Chest but only 3.1\% on RadGenome-ChestCT.
\Cref{tab:supp_counts} gives GREEN's finding counts per report.

\begin{table}[H]
  \caption{Generators on the CTRG-Chest test set.}
  \label{tab:supp_ext}
  \centering
  \footnotesize
  \setlength{\tabcolsep}{2.5pt}
  \begin{tabular}{@{}lccccc@{}}
    \toprule
    Generator & CE-F1 & GREEN & Pos-F1 & Neg-F1 & FNeg$\downarrow$ \\
    \midrule
    \method-AR & 48.9 & 42.1 & 33.5 & 65.0 & 29.6 \\
    First-observation & 47.2 & 42.2 & 36.4 & 66.4 & 25.1 \\
    \method & 46.4 & 44.1 & 33.0 & 65.4 & 27.6 \\
    \bottomrule
  \end{tabular}
\end{table}

\begin{table}[H]
  \caption{GREEN's counts of matched findings and clinically significant false and missed findings per report.}
  \label{tab:supp_counts}
  \centering
  \footnotesize
  \setlength{\tabcolsep}{2.5pt}
  \begin{tabular}{@{}llcccc@{}}
    \toprule
    Data & Generator & Matched & False & Missed & GREEN \\
    \midrule
    \multirow{2}{*}{RadGenome} & \method-AR & 5.97 & 2.43 & 2.78 & 48.3 \\
    & \method & 6.07 & 2.25 & 2.63 & 49.6 \\
    \midrule
    \multirow{2}{*}{CTRG} & \method-AR & 3.44 & 2.07 & 2.44 & 42.1 \\
    & \method & 3.61 & 1.73 & 2.52 & 44.1 \\
    \bottomrule
  \end{tabular}
\end{table}

\begin{figure*}[!t]
  \centering
  \includegraphics[width=\linewidth]{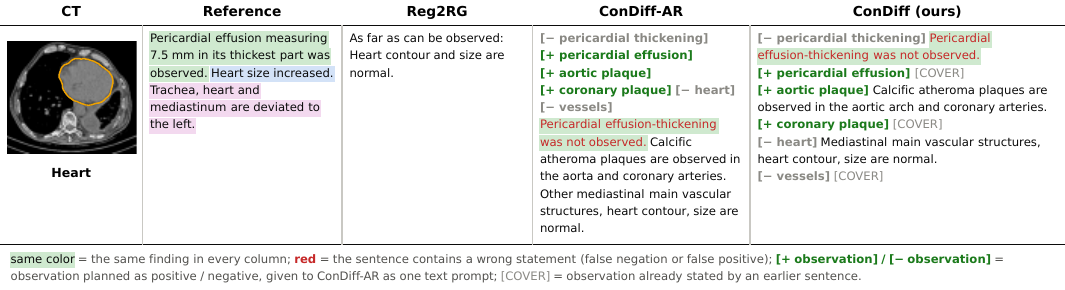}
  \caption{A failure case in the heart region section, drawn as in \cref{fig:cases}, with wrong statements in red.}
  \label{fig:supp_failure}
\end{figure*}

\section{More Examples and a Failure Case}
\label{supp:examples}
\Cref{fig:supp_cases} presents additional region sections formatted as in \cref{fig:cases}.
\Cref{fig:supp_failure} shows a failure case caused by a sentence that states multiple observations.
The segment for pericardial thickening writes the frequent sentence ``Pericardial effusion-thickening was not observed'', which simultaneously negates pericardial effusion.
The subsequent effusion segment, although planned as positive, then starts as \Cov, and the effusion is never stated.
The plan also marks the heart as normal although it is enlarged.
\method-AR makes the same error with the same sentence.

\section{Limitations and Future Work}
\label{supp:limits}
The plan of \method can include a finding only if its classifier on the frozen encoder detects it, so a finding missed by its classifier is at best left unmentioned.
On small datasets, the learned assignment can lower positive F1, as \cref{tab:supp_ext} shows for CTRG-Chest.
Future work includes joint training of the three-state classifiers with the generator and a reader study with radiologists.